\documentclass[10pt,a4paper]{article}
\usepackage[left=2.5cm,right=2.5cm,top=2.8cm,bottom=2.5cm,headheight=52pt]{geometry}
\usepackage{fontspec}
\usepackage[default]{sourcesanspro}
\newfontfamily\chinesefont{FandolSong-Regular.otf}
\newcommand{\zh}[1]{{\chinesefont #1}}
\newfontfamily\japanesefont{HaranoAjiMincho-Regular.otf}
\newcommand{\ja}[1]{{\japanesefont #1}}
\usepackage{microtype,amsmath,amssymb,booktabs,tabularx,array,multirow,graphicx,xcolor,hyperref,fancyhdr,caption,placeins,needspace}
\usepackage{longtable}
\definecolor{bilipink}{RGB}{251,114,153}
\definecolor{biliblue}{RGB}{0,140,210}
\hypersetup{colorlinks=true,linkcolor=bilipink,citecolor=bilipink,urlcolor=bilipink,breaklinks=true,pdftitle={Index-Translate: A Multilingual Translation Model Family},pdfauthor={Tianjiao Li, Mengran Yu, Chenyu Shi, Lusheng Zhang, Qisi Chen, Yanshan Zhou, Ji Qi, Jingying Liu, Yuang Feng, Ziang Cui, Tianxing Yan}}
\renewcommand{\arraystretch}{1.12}
\makeatletter
\def\@maketitle{%
  \newpage\null\vspace{-3.5em}%
  \begin{center}%
    {\color{biliblue}\hrule height 0.8pt}%
    \vspace{1.5em}%
    {\huge\bfseries\@title\par}%
    \vspace{1.5em}%
    {\color{biliblue}\hrule height 0.8pt}%
    \vspace{1.8em}%
    {\large\@author\par}%
    \vspace{0.5em}%
  \end{center}%
  \par\vspace{1.5em}}
\makeatother
\title{Index-Translate: A Multilingual Translation Model Family\\[4pt]
\large Text, Speech, Controlled Dubbing, and Long-Document Translation}
\author{%
\normalsize Tianjiao Li\thanks{Project lead and corresponding author. Email: \href{mailto:mayokaze@gmail.com}{mayokaze@gmail.com}.}, Mengran Yu, Chenyu Shi, Lusheng Zhang\\
\normalsize Qisi Chen, Yanshan Zhou, Ji Qi, Jingying Liu\\
\normalsize Yuang Feng, Ziang Cui, Tianxing Yan\\[4pt]
\normalsize Index LLM Team}
\date{}

\begin{document}
\maketitle
\thispagestyle{fancy}
\begin{abstract}
We introduce Index-Translate, a multilingual translation model family that combines a shared multilingual foundation with specialized training for general translation, instruction following, speech translation, controlled dubbing, and long-document translation. It includes three model sizes, 2B, 9B, and 35B-A3B, and supports translation in 150 languages, with multilingual instruction following. Evaluations on general translation and complex translation instructions show that Index-Translate outperforms translation models of comparable size and achieves performance comparable to 100B-scale translation models and frontier models. Image-text evaluation shows that translation remains possible without image examples in Index-Translate-specific training, although results vary by model size. Index-Echo provides end-to-end speech-to-text and speech-to-speech translation, outperforming existing end-to-end models and achieving performance comparable to frontier omni models. Index-Homura extends the family to syllable-controlled dubbing. Index-NativeLong introduces native long-document translation with a dedicated task formulation and benchmark. These capabilities support diverse translation tasks, including multilingual content production.
\end{abstract}

\begin{center}
\begin{minipage}{\linewidth}
\centering
\includegraphics[width=\linewidth]{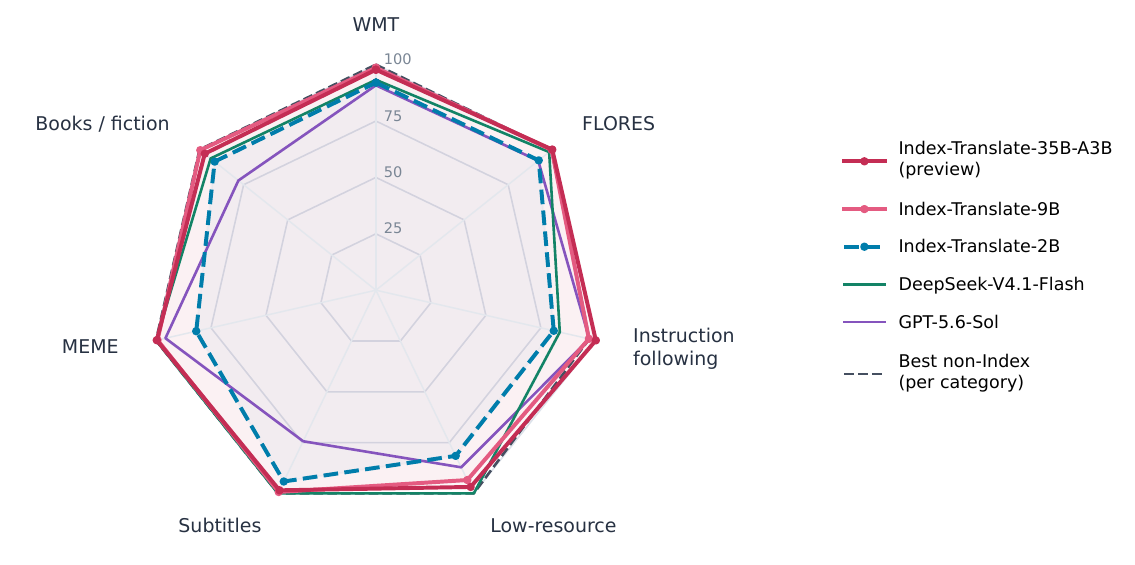}
\captionof{figure}{Seven-category text-translation comparison using the \href{https://index-translate.bilibili.com/?p=/site/home-en.html}{online demo}'s updated aggregates. Each axis is min--max normalized to 0--100 over all 14 models, including 35B-A3B (preview). Instruction following averages instTrans and IFMTBench IFscore; subtitles average OpenSubtitles and MuST-Cinema; books average OPUS-Books and GuoFeng-Webnovel. The gray dashed curve combines the best non-Index score per category.}
\label{fig:benchmark-radar}
\end{minipage}
\end{center}

\clearpage

\section{Introduction}
Multilingual translation models must preserve meaning across languages while accommodating differences in task requirements, input and output modalities, and context length. Index-Translate addresses these needs as a model family built on a shared multilingual foundation.

Content production provides concrete examples of these requirements. Structured documents must preserve fields and terminology; community expressions require cultural interpretation; dubbed utterances must fit a target length; and books must maintain identities across distant passages. Speech translation adds acoustic understanding and spoken output.

Our central question is which of these capabilities compact translation models can share and consolidate, and which benefit from dedicated training or output interfaces. Starting from Qwen3.5 base models before post-training~\cite{qwen2026qwen35}, Index-Translate addresses this question through three connected design choices:
\begin{itemize}
\setlength{\itemsep}{4pt}
\setlength{\parskip}{0pt}
\item \textbf{A shared multilingual foundation.} We combine general, monolingual, parallel, and pivot-organized text to build a translation foundation for 150 languages, studying how language coverage and data organization affect translation quality and general capability retention.
\item \textbf{Integration of complementary translation skills.} General, instruction-following, and meme experts use task-specific SFT and RL, with adversarial supervision for judge-based rewards. Parameter interpolation combines these experts, followed by targeted multi-teacher on-policy distillation (MOPD) for tasks that remain weak after merging.
\item \textbf{Dedicated training for speech, length control, and long documents.} Index-Echo connects the text foundation to speech perception and generation; Index-Homura learns explicit syllable control; and Index-NativeLong addresses native long-document translation with a dedicated task formulation and benchmark. These adaptations address different input, output, and sequence-length requirements.
\end{itemize}
Our analyses examine both gains and limits of capability sharing. Broader language coverage improves non-core translation, while expert integration exposes trade-offs among quality, instruction adherence, and cultural adaptation. Precise syllable control transfers poorly through the studied merging and distillation settings, motivating a dedicated specialist. Text-only adaptation also supports image-text translation through the inherited visual pathway, with stronger cross-modal generalization at larger scales (Section~\ref{sec:image-translation}). Table~\ref{tab:family} summarizes the model family; Appendix~\ref{app:language-inventory} lists the supported languages and their identifiers.

\begin{table}[ht]
\centering\small
\begin{tabularx}{\linewidth}{l l X}
\toprule
Model & Backbone & Task\\
\midrule
Index-Translate & 2B / 9B / 35B-A3B & Multilingual text and image-text translation, with translation instructions\\
Index-Echo S2TT & 2B / 9B & Speech-to-text translation\\
Index-Echo S2ST & 2B / 9B & Speech-to-speech translation with voice conditioning\\
Index-Homura & 2B / 9B & Translation with a specified syllable count\\
Index-NativeLong & 2B / 9B & Native long-document translation\\
\bottomrule
\end{tabularx}
\caption{Index-Translate model family and evaluated model sizes.}
\label{tab:family}
\end{table}

\FloatBarrier
\section{Training}
\begin{figure}[htbp]
\centering
\includegraphics[width=\linewidth]{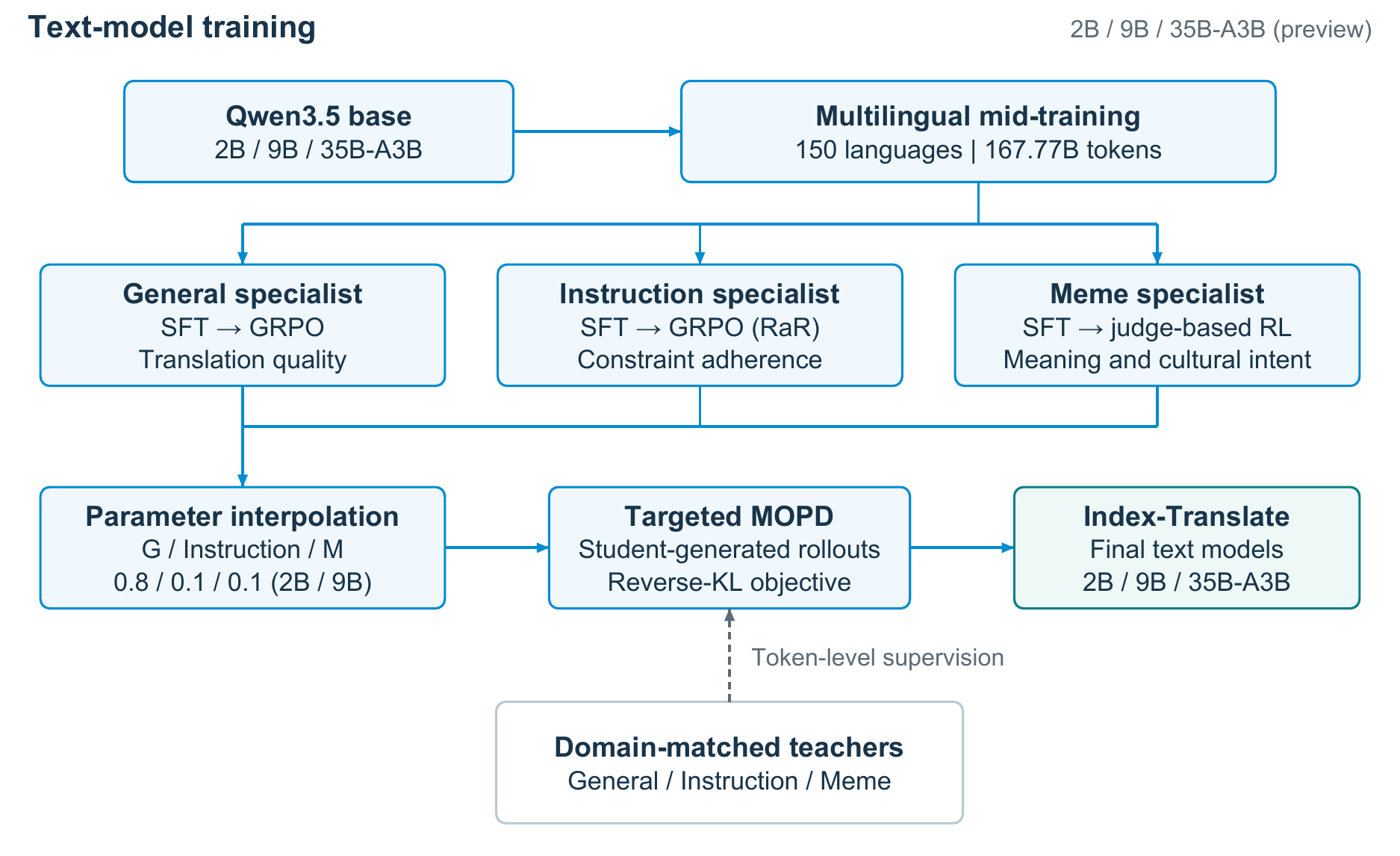}
\caption{Training overview of the text-model family, including 2B, 9B, and 35B-A3B. General, Instruction, and Meme specialists share a multilingual mid-trained initialization. Parameter interpolation is followed by targeted MOPD, using domain-matched teachers to supervise student-generated rollouts with a reverse-KL objective, to obtain the final models. The displayed interpolation weights apply to the evaluated 2B and 9B models. Judge-based rewards use RIVAL supervision as described in Section~\ref{sec:robust-judge}.}
\label{fig:training-overview}
\end{figure}

\subsection{Multilingual mid-training}
\paragraph{Data design.}
Mid-training improves cross-language alignment while retaining broad linguistic competence. We combine general pre-training replay, independently sampled monolingual text, and parallel translations.

\paragraph{Parallel and pivot organization.}
Aligning the same content across languages can improve multilingual consistency and performance~\cite{shen2025multiway}. We distinguish ordinary parallel text ($P$) from pivot-organized text ($V$). In the Qwen3-8B pilot, ordinary parallel examples align two languages directly, while pivot groups connect multiple language variants through an English anchor. Parallel degree counts the languages aligned to the same content; data-stream weights determine how often each construction is sampled.

The final recipe uses separate pivot streams for core and full language coverage. The stream labels \texttt{pivot-20} and \texttt{pivot-148} in Table~\ref{tab:midtrain-budget} exclude English and Chinese from their counts; including these two languages gives 22 core and 150 total entries in Appendix~\ref{app:language-inventory}. Ordinary parallel text enters the constant stage with general and monolingual text; pivot data dominate decay. Appendix~\ref{app:parallel-schematic} illustrates the two organizations.

\paragraph{Training recipe.}
The shared 2B, 9B, and 35B-A3B recipe follows the stable and decay phases of the warmup--stable--decay schedule~\cite{minicpm}. We use AdamW, a sequence length of 4,096, a global batch of 4,096 sequences, and a micro-batch of 128. Table~\ref{tab:midtrain-budget} gives the stage mixtures, learning rates, and token budgets. The long-document extension increases sequence length during decay (Section~\ref{sec:nailong-training}).

\begin{table}[ht]
\centering\small
\setlength{\tabcolsep}{4pt}
\begin{tabular}{llrrl}
\toprule
Stage & Sampling ratio & Steps & Tokens (B) & Learning rate\\
\midrule
Constant & general : parallel : monolingual $=1{:}1{:}1$ & 7,000 & 117.44 & $5\times10^{-5}$\\
Decay & general : pivot-20 : pivot-148 $=1{:}4{:}2$ & 3,000 & 50.33 & $5\times10^{-5}\to5\times10^{-6}$\\
\midrule
Total & & 10,000 & 167.77 &\\
\bottomrule
\end{tabular}
\caption{Base translation mid-training recipe shared by 2B, 9B, and 35B-A3B. Token budgets use 4,096 tokens per sequence and 4,096 sequences per global batch.}
\label{tab:midtrain-budget}
\end{table}

Section~\ref{sec:midtrain-analysis} analyzes recipe choices, stage progression, and language coverage.

\Needspace{8\baselineskip}
\subsection{Post-training}\label{sec:posttraining}
We develop general translation, instruction-following, and meme-translation experts through a shared SFT-to-RL workflow. Each expert uses task-specific data and supervision; Section~\ref{sec:expert-integration} describes how we combine their complementary capabilities.

\paragraph{Supervised fine-tuning.}
General translation SFT uses multilingual training data with translations regenerated by frontier models and filtered for language consistency and quality, retaining both core and low-resource directions. Instruction SFT uses 1.045 million examples across ten real-world scenarios, with constraints on structure, terminology, context, and style. It mixes instruction translation and general-capability data at a $1{:}1$ token ratio. Meme SFT adds curated examples that teach the model to resolve nonliteral meanings and preserve pragmatic intent through idiomatic paraphrase, explanation, or cultural adaptation.

\paragraph{Reinforcement learning.}
RL refines the SFT models with rewards tailored to each task. General and instruction translation use GRPO. General translation combines reference-based XCOMET-XXL with target-language validity and source-grounded adequacy judgments from Gemini 2.5 Flash. Instruction translation uses Rubric-as-Reward (RaR), with Gemini 3.1 Pro judging instruction compliance. Using $g$ for binary gates and $q$ for graded scores, the rewards are
\begin{align}
R_{\mathrm{gen}} &= g_{\mathrm{lang}}\,q_{\mathrm{XCOMET}}\,q_{\mathrm{adeq}},
\label{eq:general-reward}\\
R_{\mathrm{inst}} &= g_{\mathrm{lang}}\,g_{\mathrm{hard}}\,\bigl(q_{\mathrm{qual}}+q_{\mathrm{soft}}\bigr).
\label{eq:instruction-reward}
\end{align}
Here $g_{\mathrm{lang}}$ denotes the task-specific target-language gate, and $g_{\mathrm{hard}}$ checks required structure, terminology, and preservation of code or placeholders; a failed check sets the corresponding gate to zero. For general translation, $q_{\mathrm{XCOMET}}$ is the reference-based XCOMET-XXL score and $q_{\mathrm{adeq}}$ is the adequacy score applied as a multiplicative discount for omissions, additions, and other fidelity errors. For instruction translation, $q_{\mathrm{qual}}$ is the rubric-based translation-quality score and $q_{\mathrm{soft}}$ averages soft-constraint scores for style and contextual interpretation. Timed-subtitle constraints also assess agreement between syllable counts and segment durations.

Meme RL uses a reference-guided LLM judge to assess contextual meaning, cultural function, and pragmatic effect.

\paragraph{Adversarial judge supervision.}\label{sec:robust-judge}
We supplement judge-based rewards with our RIVAL framework~\cite{rival2025}. Training alternates reward-model updates on current policy outputs with policy optimization against the refreshed reward. Preference ranking, auxiliary quantitative supervision, and replay of earlier outputs help the evaluator track evolving failure patterns. In our earlier Qwen2.5-7B study~\cite{rival2025}, subtitle quality judged by GPT-4o rose from 3.26 to 3.68 between RIVAL iterations 0 and 1; training-curve analysis also showed reduced divergence between reward-model and external evaluations. At iteration 2 on WMT English-to-Chinese, adding quantitative supervision to preference ranking improved BLEU from 30.14 to 39.39 and COMETKiwi from 71.91 to 72.60. These results motivate using adaptive reward supervision to mitigate reward hacking. Appendix~\ref{app:judge-cases} gives representative cases from Index-Translate training.

\subsection{Expert integration}\label{sec:expert-integration}
\paragraph{Parameter interpolation.}
We first combine compatible specialists in parameter space. For the general ($G$), instruction, and meme ($M$) specialists, we use
\begin{equation}
\theta_{\mathrm{merge}}
=w_G\theta_G+w_{\mathrm{Instruction}}\theta_{\mathrm{Instruction}}+w_M\theta_M,\qquad
w_G+w_{\mathrm{Instruction}}+w_M=1.
\label{eq:merge}
\end{equation}
Parameter averaging follows the model-soup approach~\cite{wortsman2022} and produces a merged initialization for targeted MOPD. The interpolation stage for the evaluated Index-Translate-2B and 9B models uses $w_G/w_{\mathrm{Instruction}}/w_M=0.8/0.1/0.1$. We select weights using both translation quality and instruction following (Section~\ref{sec:merge-analysis}), then use post-merge evaluation to identify task types for targeted MOPD.

\Needspace{8\baselineskip}
\paragraph{Multi-teacher on-policy distillation.}
Following the MOPD framework~\cite{ma2026mopd}, we use domain-matched teachers to supervise student-generated rollouts with a reverse-KL objective. In our workflow, this serves as a targeted follow-up to parameter interpolation, addressing task types where the merged model shows weaknesses. Targeted MOPD is part of the training pipeline for the final Index-Translate models. Section~\ref{sec:merge-analysis} reports configuration studies using an earlier 2B reference merge, with general, instruction, and meme sampling ratios of $1{:}1{:}1$ or $1{:}2{:}1$, including a configuration using three 9B specialists as teachers.

\section{Model Extensions}
\subsection{Index-Echo: speech translation}
\paragraph{Speech-to-text translation.}
Index-Echo connects a Qwen3-Omni AuT audio encoder~\cite{xu2025qwen3omni} and connector to an Index-Translate 2B or 9B decoder. We train the resulting model end to end on speech translation, combining acoustic input, target-language instructions, and translated text in one sequence. This design uses the multilingual text model as the translation component while learning to interpret the source directly from audio.

\paragraph{Speech-to-speech translation.}
We initialize S2ST training from the trained 2B and 9B S2TT models. The speech-generation path replaces the CosyVoice3~\cite{du2025cosyvoice3} text-tokenizer interface with a roughly 30M-parameter Hidden2CV mapper from the translator's final hidden states to the semantic layer. We first distill the mapper with the S2TT backbone and speech generator frozen. The S2TT backbone remains frozen during subsequent differentiable reward optimization (DiffRO). Following CosyVoice3~\cite{du2025cosyvoice3}, DiffRO backpropagates a Token2Text content-consistency reward through Gumbel-Softmax speech-token samples. Source audio supplies a reference prompt and CampPlus speaker embedding for voice conditioning.

\begin{figure}[ht]
\centering
\includegraphics[width=\linewidth]{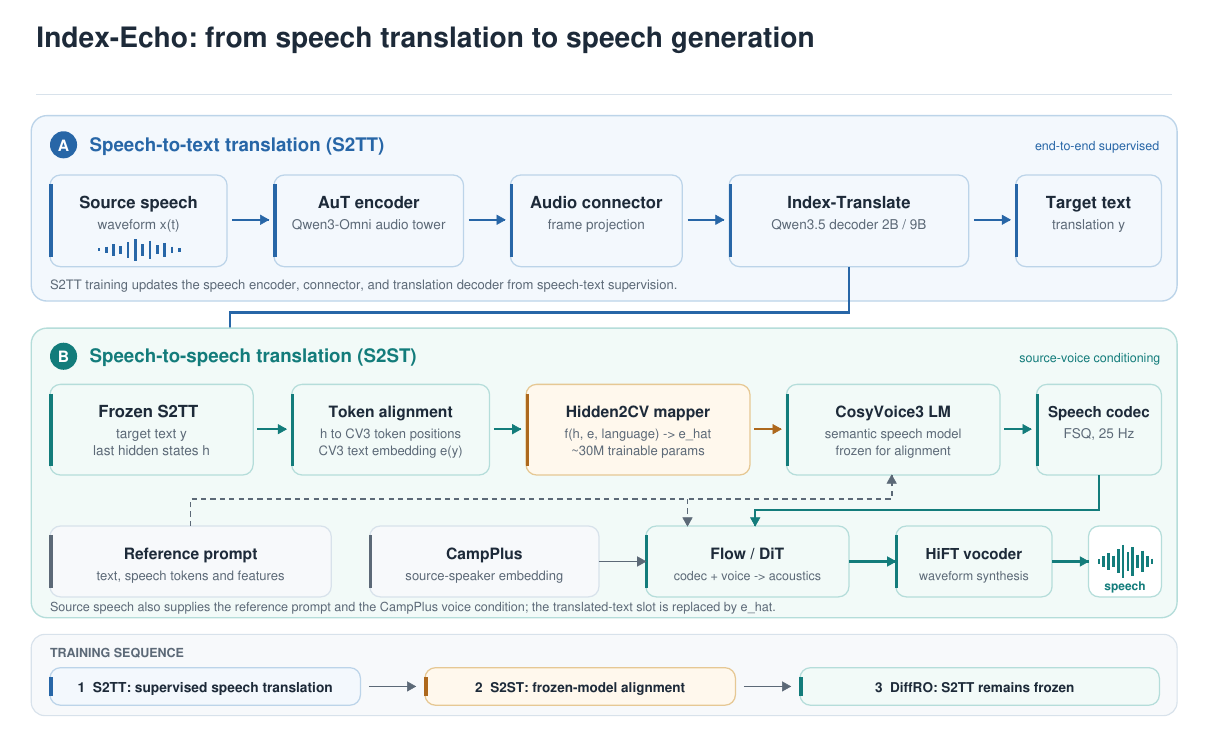}
\caption{Index-Echo speech-to-text and speech-to-speech architecture. S2ST training aligns the mapper and then applies DiffRO; the S2TT backbone remains frozen throughout.}
\label{fig:echo-arch}
\end{figure}

The resulting S2ST models cover six translation directions: Chinese to English, Spanish, and Japanese, and English to Chinese, Spanish, and Japanese. The target language is specified explicitly, and source audio provides voice conditioning.

\FloatBarrier
\subsection{Index-Homura: translation for dubbing}
Index-Homura adapts the 2B and 9B text models to translation with an explicitly specified target syllable count. Given a source text and a target count $s^\star$, the model adjusts its wording to meet the requested count while preserving meaning and natural expression.

Building on the reinforcement-learning approach of HOMURA~\cite{homura}, we use GRPO to jointly optimize translation quality and adherence to the specified syllable count. The length reward measures how closely the generated translation matches $s^\star$, enabling direct control over the syllable count of the output.

\subsection{Index-NativeLong: native long-document translation}\label{sec:nailong-training}\label{sec:nailong-analysis}
Book translation requires faithful coverage and continuity across distant passages. We study \emph{native long-document translation}, in which a model receives the full source document and produces its complete translation in a single generation. Index-NativeLong extends our multilingual text foundation to this task, combining long-context understanding with sustained output~\cite{longwriter}.

\paragraph{Why translate the full document?}
Glossary-assisted frameworks such as TranslateBooksWithLLMs~\cite{translatebooks} connect chunks through terminology lists and neighboring context, but can still lose conceptual distinctions and document-wide consistency. In our glacial-history case, 9B preserves the distinction between an \emph{ice age} (\zh{冰河时代}) and its \emph{glacials} (\zh{冰期}), while chunked Qwen3.8-Flash conflates them even with a full-source glossary. For recurring names such as \zh{王妃} and Zeitlin, native 9B maintains consistent translations where the same model with glossary-assisted chunking changes them. Appendix~\ref{app:longdoc-cases} gives the original excerpts and comparison settings.

Native generation keeps the full source and translation history together, supporting these document-level requirements without separate glossary preparation or repeated context input. We introduce NativeLongBench to evaluate this capability; Section~\ref{sec:nailong-bench} presents the data and results.

\paragraph{Training.}
We adapt the 2B and 9B models through full-parameter SFT on aligned book passages and complete target translations. The 2B model follows 128K-sequence mid-training decay and general SFT; the 9B model continues from the integrated model in Section~\ref{sec:expert-integration}. Both receive bidirectional document supervision over approximately 4K--64K Chinese-side tokens. GuoFeng ablations support long-sequence decay and show comparable overall quality for RoPE and NoPE after long-document SFT (Appendix~\ref{app:nailong-ablation}).

\FloatBarrier
\Needspace{10\baselineskip}
\section{Evaluation}
\subsection{Evaluation setup}\label{sec:eval-setup}
Unless a training stage is explicitly identified, all Index-Translate-2B and 9B evaluations use the same final released models, obtained by parameter interpolation with general, instruction, and meme weights of $0.8{:}0.1{:}0.1$, followed by targeted MOPD. We additionally evaluate Index-Translate-35B-A3B (preview). All evaluated models in the text translation and general-capability comparisons are run in non-thinking mode, with thinking explicitly disabled when supported. Speech baselines evaluated with thinking enabled are explicitly labeled in Table~\ref{tab:echo}. Text translation uses greedy decoding and a 16,384-token context window, with the WMT26 settings specified below. Training-stage studies and model extensions report their settings separately. Table~\ref{tab:text-protocol} summarizes the translation benchmarks, including FLORES-200~\cite{nllb2022}, WMT24++~\cite{deutsch2025wmt24pp}, IFMTBench~\cite{zheng2026hymt2}, and GuoFeng-Webnovel~\cite{wang2023guofeng}.

\begin{center}
\begin{minipage}{\linewidth}
\centering\small
\begin{tabularx}{\linewidth}{@{}lr>{\raggedright\arraybackslash}X@{}}
\toprule
Benchmark & Items & Coverage and split \\
\midrule
FLORES-200 & 126,000 & 420 directions among 21 languages; devtest; 300 per direction\\
FLORES low-resource pairs & 104,000 & 1,040 directions among 62 languages; devtest; 100 per direction\\
WMT24++ & 6,000 & English to 20 languages; test; 300 per direction\\
WMT26 & 4,897 & 23 directions covering 20 target languages\\
instTrans & 3,000 & Chinese to 20 languages; aligned evaluation set\\
instTrans low-resource & 2,793 & Chinese / English to low-resource languages; aligned evaluation set\\
IFMTBench & 7,064 & Translation instructions; preprocessing in Appendix~\ref{app:ifmt-preprocessing}\\
OPUS-Books & 2,700 & English to nine languages; train split; 300 per direction\\
WMT Biomedical & 600 & English paired bidirectionally with six languages; test; 50 per direction\\
MTNT & 4,013 & English paired bidirectionally with French and Japanese; test\\
MuST-Cinema & 3,803 & English to seven languages; Amara test split\\
GuoFeng-Webnovel & 25 chapters & Chinese to English; document-level evaluation\\
MEME & 3,638 & Culturally marked translation examples\\
\bottomrule
\end{tabularx}
\captionof{table}{Text translation evaluation sets. Counts refer to evaluated inputs; GuoFeng counts chapters.}
\label{tab:text-protocol}
\end{minipage}
\end{center}

Instruction evaluation uses GPT-5.6-Sol and reports translation quality and instruction following separately. IFMTBench reports instruction compliance and reference-based XCOMET-XXL; its preprocessing and scoring details are given in Appendix~\ref{app:ifmt-preprocessing}.

Reference-based translation metrics are COMET-22 (\texttt{wmt22-comet-da})~\cite{rei2022comet22} and XCOMET-XXL~\cite{guerreiro2024xcomet}. FLORES scores average directions equally; Vertical averages the five domain benchmarks in Table~\ref{tab:text-vertical}. WMT26 uses reference-free GPT-5.6-Sol judgments on a 0--100 scale, averaged over 20 target languages, with greedy decoding, zero-shot Chinese instructions, a 3,584-token context, and a 1,536-token output limit.

\subsection{Text translation}
Table~\ref{tab:text-main} summarizes text translation performance. Panel (b) includes low-resource general translation (FLORES\_minor\_pair) and instruction following (instTrans\_minor), with off-target rates reported for both. Index-Translate-35B-A3B (preview) achieves the highest FLORES COMET-22 and instTrans IFscore among the compared models (0.8794 and 0.8336), with an instTrans quality score of 0.6901 and a WMT26 judge score of 76.76. Index-Translate-9B remains competitive on general translation and instruction following. The 2B model outperforms Hy-MT2-1.8B on all metrics available for both models.

\begin{table}[ht]
\centering\scriptsize
\setlength{\tabcolsep}{2.4pt}
\textbf{(a) Main translation benchmarks}\par\smallskip
\begin{tabular}{lrrrrrrrrr}
\toprule
Model & FLORES & WMT24++ & WMT26 & \multicolumn{2}{c}{instTrans} & \multicolumn{2}{c}{IFMTBench} & Vertical & MEME\\
\cmidrule(lr){5-6}\cmidrule(lr){7-8}
 & COMET-22 & COMET-22 & Judge & Quality & IFscore & XCOMET-XXL & IFscore & mean & \\
\midrule
Index-Translate-35B-A3B (preview) & \textbf{0.8794} & 0.8586 & 76.76 & 0.6901 & \textbf{0.8336} & 0.7926 & 0.8991 & 0.8438 & 0.7405\\
Index-Translate-9B & 0.8789 & 0.8601 & 75.35 & 0.6771 & 0.8209 & 0.7957 & 0.8760 & 0.8451 & 0.7387\\
Index-Translate-2B & 0.8655 & 0.8489 & 60.26 & 0.5391 & 0.7569 & 0.7712 & 0.7584 & 0.8377 & 0.6443\\
\midrule
Hy-MT2-1.8B & 0.8522 & 0.8401 & 49.35 & 0.3181 & 0.4932 & 0.7493 & 0.7161 & 0.8314 & 0.3643\\
Hy-MT2-7B & 0.8747 & 0.8593 & 60.51 & 0.5143 & 0.6079 & 0.8049 & 0.8741 & 0.8335 & 0.5139\\
Hy-MT2-30B-A3B & 0.8787 & \textbf{0.8624} & 66.81 & 0.5725 & 0.6415 & \textbf{0.8177} & 0.9029 & \textbf{0.8459} & 0.5812\\
TranslateGemma-12B & 0.8732 & 0.8524 & 71.19 & 0.4515 & 0.3068 & 0.8023 & 0.2892 & 0.8347 & 0.4281\\
North-Small-Translate & 0.8784 & 0.8578 & 68.37 & 0.5697 & 0.5294 & 0.7657 & 0.8635 & 0.8357 & 0.6836\\
\midrule
Qwen3.5-2B & 0.6983 & 0.6933 & 32.11 & 0.0999 & 0.2431 & 0.6197 & 0.3836 & 0.7557 & 0.2062\\
Qwen3.5-9B & 0.8316 & 0.8073 & 60.31 & 0.2467 & 0.0609 & 0.7341 & 0.5980 & 0.8199 & 0.5728\\
Qwen3.5-35B-A3B & 0.8570 & 0.8290 & 71.33 & 0.3690 & 0.5204 & 0.7589 & 0.7822 & 0.8267 & 0.6447\\
\midrule
DeepSeek-V4.1-Flash & 0.8762 & 0.8510 & 83.55 & 0.6068 & 0.6374 & 0.7817 & 0.9090 & 0.8432 & \textbf{0.7424}\\
GPT-5.6-Sol & 0.8650 & 0.8469 & \textbf{89.10} & \textbf{0.6902} & 0.7624 & 0.7946 & \textbf{0.9367} & 0.8311 & 0.7194\\
Gemini 3.5 Flash Lite & 0.8750 & 0.8497 & 79.52 & 0.6068 & 0.6374 & 0.7764 & 0.8854 & 0.8131 & 0.7034\\
\bottomrule
\end{tabular}
\par\smallskip
\textbf{(b) Low-resource translation and instruction following}\par\smallskip
\begin{tabular}{lrrrrrr}
\toprule
Model & \multicolumn{3}{c}{FLORES\_minor\_pair} & \multicolumn{3}{c}{instTrans\_minor}\\
\cmidrule(lr){2-4}\cmidrule(lr){5-7}
 & COMET-22 $\uparrow$ & XCOMET-XXL $\uparrow$ & off-target $\downarrow$ & Quality $\uparrow$ & IFscore $\uparrow$ & off-target $\downarrow$\\
\midrule
Index-Translate-35B-A3B (preview) & 0.8168 & 0.7164 & 2.4\% & 0.5151 & 0.7715 & 4.05\%\\
Index-Translate-9B & 0.7992 & 0.6805 & 4.0\% & 0.5222 & \textbf{0.7725} & \textbf{3.47\%}\\
Index-Translate-2B & 0.7377 & 0.4817 & 4.2\% & 0.3050 & 0.6586 & 3.97\%\\
\midrule
Hy-MT2-1.8B & 0.3163 & 0.2014 & 54.2\% & 0.0412 & 0.1313 & 65.27\%\\
Hy-MT2-7B & 0.4626 & 0.3334 & 35.7\% & 0.1121 & 0.2405 & 45.40\%\\
Hy-MT2-30B-A3B & 0.6746 & 0.5359 & 14.5\% & 0.2246 & 0.4449 & 15.47\%\\
TranslateGemma-12B & 0.8021 & 0.6217 & 1.7\% & 0.2412 & 0.2682 & 5.37\%\\
North-Small-Translate & 0.6855 & 0.4919 & 13.2\% & 0.2436 & 0.2902 & 26.89\%\\
\midrule
Qwen3.5-2B & 0.3716 & 0.1852 & 39.1\% & 0.0069 & 0.0549 & 73.54\%\\
Qwen3.5-9B & 0.6774 & 0.4502 & 11.2\% & 0.1318 & 0.3532 & 18.51\%\\
Qwen3.5-35B-A3B & 0.7854 & 0.6257 & 3.5\% & 0.2874 & 0.4458 & 11.60\%\\
\midrule
DeepSeek-V4.1-Flash & \textbf{0.8333} & \textbf{0.7297} & \textbf{1.3\%} & 0.4793 & 0.5854 & 5.73\%\\
GPT-5.6-Sol & 0.7669 & 0.6918 & 10.4\% & \textbf{0.5757} & 0.6866 & 7.73\%\\
Gemini 3.5 Flash Lite & 0.8122 & 0.6995 & 3.3\% & 0.3927 & 0.5584 & 12.18\%\\
\bottomrule
\end{tabular}
\caption{Text translation results for Index-Translate and comparison systems. Instruction judging uses GPT-5.6-Sol; quality and IFscore are separate metrics. WMT26 judge scores use a 0--100 scale, off-target rates are percentages, and other scores use 0--1. Vertical is the equal-weight mean of five domain scores. North-Small-Translate denotes the 218B-A25B model, and TranslateGemma denotes \texttt{translategemma-12b-it}. Panel (b) reports low-resource general translation and instruction following. Bold marks the best value in each column: lower is better for off-target rates, and higher is better for all other metrics.}
\label{tab:text-main}
\end{table}

\paragraph{Low-resource translation and instruction following.}
FLORES\_minor\_pair covers 104,000 inputs across 1,040 directions among 62 languages; instTrans\_minor adds 2,793 instruction-translation tasks from Chinese or English to low-resource languages. Table~\ref{tab:text-main}(b) reports translation quality, instruction following, and off-target rates (the percentage of outputs not in the target language; lower is better). Among the three Index-Translate models, 35B-A3B (preview) achieves the highest low-resource FLORES COMET-22 and XCOMET-XXL scores (0.8168 and 0.7164), with a 2.4\% off-target rate. On instTrans\_minor, Index-Translate-9B achieves the highest IFscore (0.7725) and lowest off-target rate (3.47\%) among all compared models; its quality score is 0.5222.

\paragraph{Vertical-domain translation.}
Table~\ref{tab:text-vertical} expands the five-domain Vertical mean. Index-Translate-9B leads the compared models on MuST-Cinema.

\begin{table}[ht]
\centering\scriptsize
\setlength{\tabcolsep}{4pt}
\begin{tabular}{lrrrrr}
\toprule
Model & OPUS-Books & Biomedical & MTNT & MuST-Cinema & GuoFeng \\
\midrule
Index-Translate-35B-A3B (preview) & 0.8013 & 0.8681 & 0.8244 & 0.8962 & 0.8291\\
Index-Translate-9B & 0.8010 & 0.8674 & 0.8246 & \textbf{0.8979} & \textbf{0.8345}\\
Index-Translate-2B & 0.7845 & 0.8635 & 0.8132 & 0.8930 & \textbf{0.8345}\\
\midrule
Hy-MT2-1.8B & 0.7773 & 0.8592 & 0.8080 & 0.8838 & 0.8286\\
Hy-MT2-7B & 0.7954 & 0.8671 & 0.8245 & 0.8929 & 0.7876\\
Hy-MT2-30B-A3B & \textbf{0.8054} & \textbf{0.8686} & 0.8279 & 0.8966 & 0.8310\\
TranslateGemma-12B & 0.8006 & 0.8673 & 0.8107 & 0.8886 & 0.8062\\
North-Small-Translate & 0.8051 & 0.8667 & 0.8223 & 0.8936 & 0.7907\\
\midrule
Qwen3.5-2B & 0.6375 & 0.8296 & 0.7203 & 0.7951 & 0.7959\\
Qwen3.5-9B & 0.7485 & 0.8618 & 0.7917 & 0.8728 & 0.8247\\
Qwen3.5-35B-A3B & 0.7745 & 0.8645 & 0.8063 & 0.8839 & 0.8044\\
\midrule
DeepSeek-V4.1-Flash & 0.7962 & 0.8674 & \textbf{0.8290} & 0.8960 & 0.8274\\
GPT-5.6-Sol & 0.7790 & 0.8676 & 0.8255 & 0.8705 & 0.8127\\
Gemini 3.5 Flash Lite & 0.7845 & 0.8675 & 0.8238 & 0.8922 & 0.6977\\
\bottomrule
\end{tabular}
\caption{Vertical-domain translation results (higher is better): COMET-22 for the four sentence-level benchmarks and document-COMET for GuoFeng-Webnovel. Biomedical denotes the WMT Biomedical Translation Task. The equal-weight mean of these five columns is the Vertical score in Table~\ref{tab:text-main}. Bold marks the best score in each column.}
\label{tab:text-vertical}
\end{table}

\Needspace{8\baselineskip}
\paragraph{instTrans benchmark.}\label{sec:insttrans-benchmark}
We introduce instTrans to evaluate translation under task-specific instructions across subtitles, social content, articles, academic documents, books, and fiction. The evaluated core set contains 3,000 examples from Chinese to 20 target languages; a low-resource extension adds 2,793 examples from Chinese or English to low-resource languages. The benchmark covers ten constraint types, comprising five hard checks and five graded soft constraints. These include format and terminology preservation, subtitle syllable alignment, style consistency, cross-sentence terminology, coreference resolution, and context disambiguation. Multiple constraints can apply to the same example.

We report translation quality and IFscore separately. Quality uses three GPT-5.6-Sol judge calls per example on a 0/0.5/1 scale. IFscore gates the mean soft-constraint score by hard-constraint checks, assigning zero if any required hard check fails, then averages over examples. Valid quality judgments cover 2,961 / 2,960 core examples and 2,777 / 2,776 low-resource examples for 2B / 9B; means use valid judgments. Table~\ref{tab:instruction-context} breaks down selected constraints, and Appendix~\ref{app:translation-cases} provides representative translation cases.

\Needspace{8\baselineskip}
\paragraph{Meme benchmark.}\label{sec:meme-benchmark}
MEME evaluates Chinese-to-English translation of culturally marked user-generated content, with 3,638 cases covering 703 terms and 857 disambiguated senses. Community experts validate sense assignments, and bilingual experts annotate translation strategies and quality-tiered references. Gemini-2.5-Flash receives each term's definition, usage examples, and references, assigning 1 for an accurate and idiomatic translation, 0.5 for a flawed but usable translation, and 0 for an unusable translation. Appendix~\ref{app:meme-cases} illustrates how the models recover playful spellings, community aliases, and laughter expressions.

\FloatBarrier
\Needspace{9\baselineskip}
\subsection{Image-text translation}\label{sec:image-translation}
Index-Translate inherits its visual pathway from the Qwen3.5 base models before post-training~\cite{qwen2026qwen35}; its own mid- and post-training use no images. We evaluate image-text translation on three probes: the public MMTIT-Bench~\cite{li2026mmtitbench} release (1,243 images in 14 source languages, translated into English and Chinese), 1,200 plain-background images rendered from FLORES\_minor\_pair text (60 source languages, two targets), and Vistra~\cite{salesky2024vistra} (763 natural images, four targets). For Vistra, we join segment references in reading order to score whole-image output; this is an adapted protocol rather than an official benchmark score.

Each model receives only the image and a zero-shot translation instruction; source transcriptions and references are withheld from inference and used only for scoring. We compare Index-Translate 2B, 9B, and 35B-A3B (preview) with corresponding post-trained Qwen3.5 models. All use non-thinking decoding with temperature zero, top-p one, a 24,576-token context window, and a 4,096-token output limit. We report reference-based COMET-22~\cite{rei2022comet22}; the metric evaluates translated text rather than visual layout or text replacement in the image.

\begin{table}[ht]
\centering\small
\setlength{\tabcolsep}{4pt}
\begin{tabular}{llrrrl}
\toprule
Probe & Size & Index & Qwen3.5 & $\Delta$ & Paired 95\% CI\\
\midrule
MMTIT & 2B & 0.5204 & 0.6826 & $-0.1622$ & $[-0.1705,-0.1538]$\\
MMTIT & 9B & 0.7347 & 0.7667 & $-0.0320$ & $[-0.0378,-0.0262]$\\
MMTIT & 35B-A3B & 0.7845 & 0.7807 & $+0.0038$ & $[-0.0005,+0.0079]$\\
\midrule
FLORES-rendered & 2B & 0.5104 & 0.5221 & $-0.0117$ & $[-0.0223,-0.0012]$\\
FLORES-rendered & 9B & 0.6780 & 0.6714 & $+0.0066$ & $[-0.0018,+0.0147]$\\
FLORES-rendered & 35B-A3B & 0.7050 & 0.6926 & $+0.0124$ & $[+0.0036,+0.0217]$\\
\midrule
Vistra (adapted) & 2B & 0.5505 & 0.6459 & $-0.0955$ & $[-0.1040,-0.0867]$\\
Vistra (adapted) & 9B & 0.6953 & 0.7143 & $-0.0190$ & $[-0.0265,-0.0122]$\\
Vistra (adapted) & 35B-A3B & 0.7168 & 0.7031 & $+0.0137$ & $[+0.0083,+0.0189]$\\
\bottomrule
\end{tabular}
\caption{Image-text translation COMET-22. Qwen3.5 columns use post-trained models; the Index 35B-A3B checkpoint is a preview. $\Delta$ is Index minus Qwen3.5. Paired bootstrap intervals cluster by image for MMTIT and Vistra and by source language for synthetic FLORES-rendered images. Vistra uses an adapted whole-image protocol; neither supplementary probe reports an official source-benchmark score.}
\label{tab:image-translation}
\end{table}

Despite using no images during mid- or post-training, Index-Translate achieves image-text translation performance comparable to corresponding post-trained Qwen3.5 models at all three scales. Larger models show stronger cross-modal generalization.

\FloatBarrier
\Needspace{12\baselineskip}
\subsection{General capabilities}\label{sec:general-capabilities}
We evaluate general knowledge and multilingual question answering with C-Eval~\cite{huang2023ceval}, GPQA-Diamond~\cite{rein2023gpqa}, INCLUDE~\cite{romanou2024include}, and MMMLU~\cite{openai2024mmmlu}. All models use the same evaluated subsets: 12,342 labeled C-Eval test questions across 52 subjects, 198 GPQA-Diamond questions, 22,639 INCLUDE questions in 44 languages, and 4,200 MMMLU questions sampled as 300 per language across 14 languages (seed 42). Scores are answer accuracy, with sample-level averaging for the multilingual sets.

Table~\ref{tab:general-capabilities} reports the results. Index-Translate-9B exceeds Hy-MT2-7B on all four benchmarks, while remaining below Qwen3.5-9B.

\begin{table}[ht]
\centering\small
\setlength{\tabcolsep}{6pt}
\begin{tabular}{lrrrr}
\toprule
Model & C-Eval & GPQA-Diamond & INCLUDE & MMMLU\\
\midrule
Index-Translate-35B-A3B (preview) & 77.6 & 49.1 & 68.8 & 71.9\\
Index-Translate-9B & 69.6 & 36.3 & 63.1 & 66.6\\
Index-Translate-2B & 42.6 & 27.4 & 39.1 & 42.7\\
\midrule
Hy-MT2-1.8B & 46.0 & 25.7 & 37.6 & 40.1\\
Hy-MT2-7B & 57.4 & 30.5 & 50.3 & 52.3\\
Hy-MT2-30B-A3B & 78.0 & 48.9 & 67.4 & 70.3\\
North-Small-Translate & 68.6 & 37.9 & 63.4 & 67.0\\
\midrule
Qwen3.5-2B & 62.8 & 44.0 & 40.5 & 47.9\\
Qwen3.5-9B & 86.6 & 75.8 & 70.2 & 75.9\\
Qwen3.5-35B-A3B & 89.8 & \textbf{78.8} & 76.7 & 83.0\\
\midrule
DeepSeek-V4.1-Flash & \textbf{90.7} & 75.6 & 82.0 & 83.6\\
GPT-5.6-Sol & 87.5 & 72.5 & 84.5 & \textbf{85.8}\\
Gemini 3.5 Flash Lite & 88.2 & 74.0 & \textbf{84.7} & 85.0\\
\bottomrule
\end{tabular}
\caption{General-capability accuracy (\%) on matched evaluation subsets. Index-Translate uses the same model versions as the text-translation results. INCLUDE and MMMLU are averaged over examples; MMMLU samples 300 questions per language. Bold marks the highest score per column.}
\label{tab:general-capabilities}
\end{table}

\FloatBarrier
\subsection{Speech translation}
\paragraph{Speech-to-text translation.}
The in-house video translation set contains 140 windows of 50--60 seconds, with 20 per direction: Chinese to English, Japanese, Korean, Spanish, Portuguese, and Arabic, plus English to Chinese. MT judge averages Gemini-3.1-Pro's per-line translation quality ratings on a 0/0.5/1 scale. ASR error is the median error rate between the model's transcription of the source audio and the reference transcript. Timestamp MAE is the mean absolute difference, in seconds, between predicted and reference sentence start times.

Index-Echo-2B and 9B achieve the two lowest ASR and timestamp errors (Table~\ref{tab:echo}). The 9B translation score is 0.857, between Qwen3.8-Omni-Flash and Gemini-3.1-Pro with thinking. Qwen3.8-LiveTranslate is evaluated in streaming mode.

\begin{table}[!ht]
\centering\small
\begin{tabular}{lrrr}
\toprule
Model & MT judge $\uparrow$ & ASR error (median) $\downarrow$ & Start-time MAE (s) $\downarrow$\\
\midrule
Qwen3.8-Omni-Flash & \textbf{0.887} & 0.112 & 1.004\\
Index-Echo-9B & 0.857 & 0.102 & 0.488\\
Gemini-3.1-Pro (thinking) & 0.849 & 0.169 & 1.824\\
Qwen3.8-LiveTranslate & 0.833 & 0.186 & --\\
Index-Echo-2B & 0.825 & \textbf{0.088} & \textbf{0.395}\\
Gemini-2.5-Flash (thinking) & 0.817 & 0.201 & 2.402\\
Gemini-2.5-Flash & 0.757 & 0.479 & 1.746\\
Qwen3-Omni & 0.700 & 0.139 & 1.145\\
FireRed Audio & 0.448 & 0.128 & 0.765\\
SeamlessM4T-v2 & 0.060 & 1.000 & --\\
\bottomrule
\end{tabular}
\caption{S2TT comparison on the in-house video translation test set. The (thinking) label identifies baselines evaluated with thinking enabled. Dashes denote unavailable timestamp outputs. SeamlessM4T-v2 lacks line-level timestamp instructions and receives unchunked approximately 57-second inputs, beyond its approximately 30-second effective range.}
\label{tab:echo}
\end{table}

\paragraph{Speech-to-speech translation.}
We compare E2E dubbing with Index-Echo-S2TT + CosyVoice3 on in-house video dubbing data across six directions and two model sizes (Table~\ref{tab:echo-s2st-matched}). Each pair shares the same frozen Index-Echo-S2TT model and text translation, judged by Gemini-3.1-Pro. Chinese-source tests contain 100 examples for English and 200 each for Spanish and Japanese. English-source tests contain 200 examples each for Chinese, Spanish, and Japanese, with 146--198 retained after filtering. Content error is reported as mean / median, using WER for English and Spanish, normalized CER for Chinese, and katakana CER for Japanese. Speaker similarity is cosine similarity against the source speaker.

E2E lowers mean content error in 8 of 12 size--direction pairs; Chinese-to-English favors the pipeline. Speaker similarity differences are small, with a 0.021 E2E gain for Chinese-to-English and absolute differences within 0.01 elsewhere. In an earlier Chinese-source evaluation, Index-Echo-2B also exceeds SeamlessM4T-v2 in MT judge score: 0.905 versus 0.370 for English, 0.823 versus 0.343 for Spanish, and 0.838 versus 0.310 for Japanese.

\begin{table}[!ht]
\centering\footnotesize
\setlength{\tabcolsep}{4pt}
\renewcommand{\arraystretch}{1.17}
\begin{tabular}{@{}llrrrrr@{}}
\toprule
 & & & \multicolumn{2}{c}{Content error $\downarrow$ (mean / median)} & \multicolumn{2}{c}{Spk. $\uparrow$}\\
Direction / metric & Size & Text judge $\uparrow$ & Pipeline & E2E & Pipeline & E2E\\
\midrule
zh$\to$en / WER & 2B & 0.830 & \textbf{0.0450} / 0.000 & 0.0709 / 0.029 & 0.722 & 0.743\\
zh$\to$en / WER & 9B & 0.840 & \textbf{0.0477} / 0.000 & 0.0615 / 0.000 & 0.718 & 0.739\\
\addlinespace[2pt]
zh$\to$es / WER & 2B & 0.728 & 0.0811 / 0.070 & \textbf{0.0704} / 0.051 & 0.747 & 0.743\\
zh$\to$es / WER & 9B & 0.793 & 0.0916 / 0.071 & \textbf{0.0773} / 0.057 & 0.746 & 0.746\\
\addlinespace[2pt]
zh$\to$ja / Kata CER & 2B & 0.760 & 0.0575 / 0.034 & \textbf{0.0493} / 0.027 & 0.777 & 0.780\\
zh$\to$ja / Kata CER & 9B & 0.815 & 0.0552 / 0.029 & \textbf{0.0436} / 0.021 & 0.778 & 0.782\\
\addlinespace[2pt]
en$\to$zh / CER & 2B & 0.825 & 0.0614 / 0.000 & \textbf{0.0372} / 0.000 & 0.630 & 0.623\\
en$\to$zh / CER & 9B & 0.800 & \textbf{0.0463} / 0.000 & 0.0485 / 0.000 & 0.632 & 0.623\\
\addlinespace[2pt]
en$\to$es / WER & 2B & 0.805 & 0.0867 / 0.070 & \textbf{0.0702} / 0.000 & 0.732 & 0.735\\
en$\to$es / WER & 9B & 0.820 & 0.0819 / 0.000 & \textbf{0.0660} / 0.000 & 0.736 & 0.729\\
\addlinespace[2pt]
en$\to$ja / Kata CER & 2B & 0.790 & \textbf{0.0342} / 0.000 & 0.0345 / 0.000 & 0.707 & 0.698\\
en$\to$ja / Kata CER & 9B & 0.803 & 0.0347 / 0.000 & \textbf{0.0337} / 0.000 & 0.705 & 0.710\\
\bottomrule
\end{tabular}
\caption{Matched S2ST comparison on in-house video dubbing test data at 2B and 9B. Pipeline denotes Index-Echo-S2TT + CosyVoice3. Both systems share the same frozen Index-Echo-S2TT model and text translation. Bold marks the lower mean content error within each row. Spk. is source-speaker cosine similarity.}
\label{tab:echo-s2st-matched}
\end{table}

\FloatBarrier

\Needspace{10\baselineskip}
\subsection{Syllable-controlled translation}
SandGlass contains 300 subtitle sentences, with 60 each from animation, film and television, travel, gaming, and knowledge, and at most ten per video. Four target languages and three target syllable counts yield 3,600 cases per model. Central counts multiply subtitle duration by speaking rates of 6.19/7.84/7.82/8.4 syllables per second for English/Japanese/Spanish/Arabic; short and long targets scale these by 0.75 and 1.25. Local models use greedy decoding with a 512-token output limit; API baselines use their recorded settings. Gemini-2.5-Flash judges quality without a reference on a 0/0.5/1 scale, with the length constraint in the prompt. We report relative syllable deviation, rates within one syllable or 10\% of the target, and output-versus-target regression slope across 1,200 three-budget groups (ideal: 1).

Index-Homura-9B and 2B achieve within-10\% rates of 81.92\% and 63.08\%, exceeding the external baselines while reducing mean deviation (Table~\ref{tab:homura-expanded}). DeepSeek-V4-Flash and GPT-5.6-Sol score higher on translation quality. Table~\ref{tab:homura-language} gives the language breakdown; Section~\ref{sec:instruction-analysis} examines the quality--control trade-off. Appendix~\ref{app:homura-cases} illustrates responses to different target counts.

\begin{table}[!ht]
\centering\footnotesize
\setlength{\tabcolsep}{4pt}
\begin{tabular}{lrrrrr}
\toprule
Model & Quality $\uparrow$ & Mean $|d|$ $\downarrow$ & Within $\pm1$ $\uparrow$ & Within 10\% $\uparrow$ & Slope $\approx1$\\
\midrule
Index-Homura-9B (RL) & 0.7863 & \textbf{0.0693} & \textbf{74.42\%} & \textbf{81.92\%} & \textbf{0.968}\\
Index-Homura-9B (SFT) & 0.8581 & 0.1752 & 41.39\% & 45.75\% & 0.680\\
Index-Homura-2B (RL) & 0.7615 & 0.0962 & 54.39\% & 63.08\% & 0.947\\
Index-Homura-2B (SFT) & 0.7965 & 0.1594 & 37.89\% & 43.06\% & 0.672\\
\midrule
GPT-5.6-Sol (low) & 0.8715 & 0.3223 & 42.53\% & 47.64\% & 0.891\\
DeepSeek-V4-Flash (no thinking) & \textbf{0.8742} & 0.3368 & 16.58\% & 16.36\% & 0.153\\
Hy-MT2-7B & 0.8560 & 0.3174 & 18.14\% & 18.86\% & 0.195\\
Hy-MT2-30B-A3B & 0.7554 & 0.2743 & 20.97\% & 22.39\% & 0.340\\
Hy-MT2-1.8B & 0.5904 & 0.7080 & 14.58\% & 16.72\% & 0.222\\
Hunyuan-MT-7B & 0.7188 & 1.0579 & 12.31\% & 15.11\% & 0.031\\
Qwen3.5-9B & 0.7183 & 0.3247 & 15.61\% & 16.39\% & 0.430\\
Qwen3.5-35B-A3B & 0.7863 & 0.3460 & 13.31\% & 13.25\% & 0.249\\
Qwen3.5-2B & 0.4917 & 0.5151 & 12.50\% & 12.25\% & 0.036\\
\bottomrule
\end{tabular}
\caption{Expanded SandGlass evaluation on 3,600 cases per model. $d$ is relative syllable-count deviation. Bold marks the best value among the displayed models (slope closest to 1). GPT-5.6-Sol uses the recorded gateway configuration, which may inject additional system prompts.}
\label{tab:homura-expanded}
\end{table}

\begin{table}[ht]
\centering\small
\begin{tabular}{lrrr}
\toprule
Target language & Cases & Quality $\uparrow$ & Within 10\% $\uparrow$\\
\midrule
English & 900 & 0.7950 & 78.67\%\\
Japanese & 900 & 0.8300 & 86.67\%\\
Arabic & 900 & 0.7661 & 79.89\%\\
Spanish & 900 & 0.7539 & 82.44\%\\
\bottomrule
\end{tabular}
\caption{Index-Homura-9B by target language on SandGlass. Each language includes 300 sentences at three length targets.}
\label{tab:homura-language}
\end{table}

\FloatBarrier
\subsection{Native long-document translation}\label{sec:nailong-bench}
\paragraph{Book data and benchmark.}
NativeLongBench evaluates native long-document translation on GuoFeng~\cite{wang2023guofeng}, BWB~\cite{jiang2022bwb}, and General Books. We select and align bilingual book material from these corpora. GuoFeng and BWB contain web fiction originally written in Chinese, with English translations. General Books adds English originals across literature, history, social science, science and technology, paired with model-generated Chinese translations. We preserve chapter order, align bilingual passages, and filter incomplete or mismatched pairs. Table~\ref{tab:nailong-data} summarizes the books used in this study. Japanese originals paired with model-generated Chinese translations provide additional training supervision.

All three corpora are divided by work into training, validation, and test sets. GuoFeng test works are selected from its public training corpus because the official test set is not public. We also exclude identified copies of test works in other corpora from training. The benchmark draws continuous passages from held-out books, evaluating Chinese to English on GuoFeng/BWB and English to Chinese on General Books. This GuoFeng evaluation is distinct from the chapter-level test in Table~\ref{tab:text-protocol}.

\begin{table}[!ht]
\centering\footnotesize
\setlength{\tabcolsep}{5pt}
\begin{tabular}{@{}llrrrrrr@{}}
\toprule
Corpus & Original & Books & Train & Valid. & Test & Mean (K) & P25--P75 (K)\\
\midrule
GuoFeng & Chinese & 140 & 130 & 4 & 6 & 171.6 & 56.7--87.7\\
BWB & Chinese & 337 & 327 & 4 & 6 & 777.2 & 277.0--1205.4\\
General Books & English & 306 & 292 & 3 & 11 & 103.2 & 72.3--127.5\\
\bottomrule
\end{tabular}
\caption{Books used in this study. Lengths describe the available original-language text per training book (K = 1,024 tokens); P25--P75 denotes the middle 50\% of the length distribution.}
\label{tab:nailong-data}
\end{table}

We plan to make the shareable portions of NativeLongBench publicly available.

\paragraph{Evaluation.}
The main comparison evaluates all models in the native setting: each receives the complete test document and is asked to translate it in a single generation. All test length groups use Chinese-side token counts: Chinese source text for GuoFeng/BWB and Chinese reference translations for General Books. The five groups span 4K--64K; the General Books 64K group corresponds to approximately 68K--82K English input tokens. These fixed evaluation sets have also been used during model development. Index-NativeLong uses greedy decoding with a 256K-token context window.

Gemini 3.8 Flash is excluded from the main comparison because its 65,536-token output limit~\cite{gemini38flash} falls below the 128K output budget reserved for the longest test group. North's long-input results use an experimental extension beyond its supported 16K input/output range. Some GLM requests are refused; the affected results are marked in Table~\ref{tab:nailong-frontier}.

\paragraph{Scoring.}
We use SEGALE~\cite{wang2025segale} to align the generated translation with the source before applying COMET~\cite{rei2020comet}. The output is segmented into sentences, and semantic alignment matches contiguous source and output blocks. Each matched output block is paired with its source text and the corresponding reference translation, allowing sentence-level COMET to evaluate long documents despite differing sentence boundaries. Unmatched blocks receive zero. We average block scores within each document, then document scores within each length group, and finally weight the five groups equally per corpus.

\begin{figure}[!htbp]
\centering
\includegraphics[width=\linewidth]{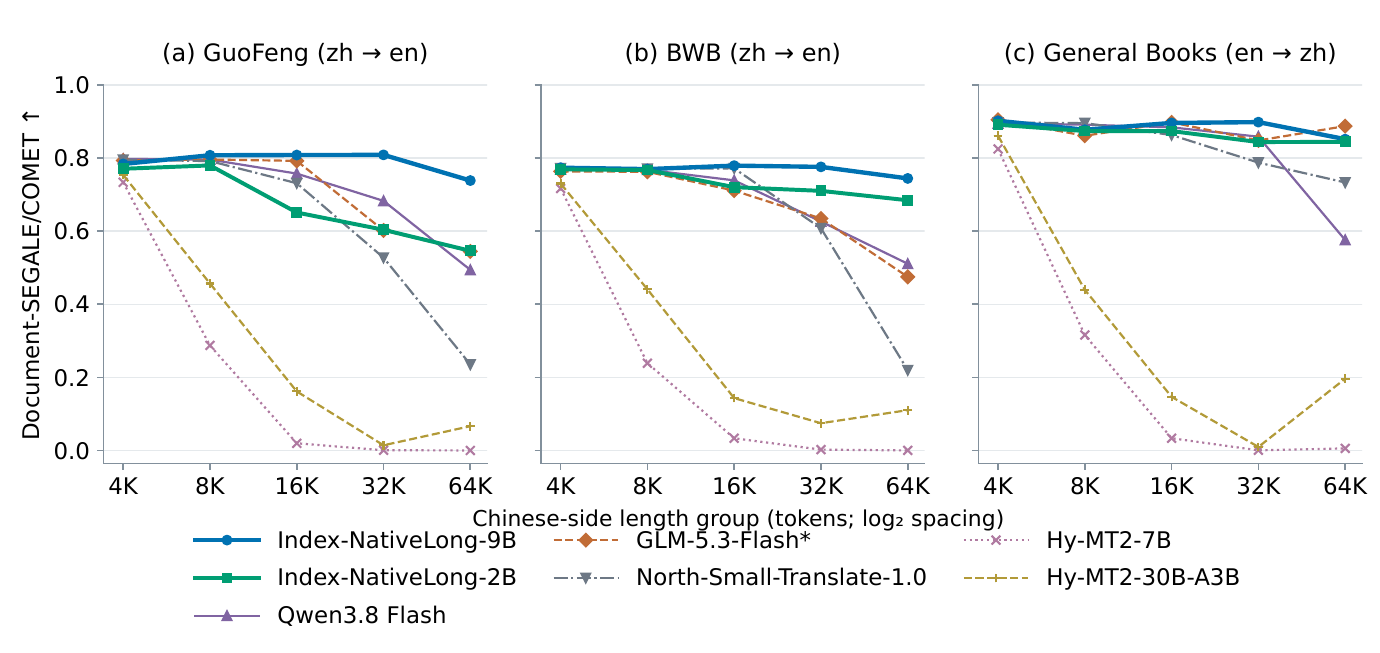}
\caption{Native long-document translation across Chinese-side length groups, reproducing the point estimates in Table~\ref{tab:nailong-frontier}. Higher document-SEGALE/COMET is better; unmatched blocks receive zero, so the score reflects coverage as well as translation quality. $^{*}$GLM's General Books scores average non-refused samples. North uses the experimental context extension described in Section~\ref{sec:nailong-bench}.}
\label{fig:native-long-quality}
\end{figure}

\paragraph{Results.}
Index-NativeLong supports the full tested length range and achieves strong quality under native generation. The 9B model has the highest five-group means in Table~\ref{tab:nailong-frontier} on GuoFeng, BWB, and General Books: 0.7891, 0.7683, and 0.8848. Several models score similarly on short passages, but their quality diverges as length increases (Figure~\ref{fig:native-long-quality}). On GuoFeng, Qwen3.8 Flash falls from 0.7974 at 4K to 0.4939 at 64K, while Index-NativeLong-9B changes from 0.7836 to 0.7382. The 2B model also exceeds the external models at 32K and 64K on BWB. These results show that compact models adapted through document training can outperform existing translation and frontier models on native long-document translation.

The outputs reveal different failure modes. Hy-MT2 sometimes stops after a short fragment; North can degenerate into repetition without stopping on long inputs; Flash can leave sentences unfinished or retain source-language text. Some 2B outputs repeat passages, possibly reflecting the limits of model capacity during sustained generation.

\begin{table}[!ht]
\centering\footnotesize
\setlength{\tabcolsep}{3.5pt}
\renewcommand{\arraystretch}{1.15}
\begin{tabular}{@{}lrrrrrr@{}}
\toprule
Model & 4K & 8K & 16K & 32K & 64K & Mean\\
\midrule
\multicolumn{7}{@{}l}{\textit{GuoFeng (zh$\to$en)}}\\[2pt]
Qwen3.8 Flash & $\mathbf{0.7974}$ & $0.7956$ & $0.7572$ & $0.6826$ & $0.4939$ & $0.7053$\\
GLM-5.3-Flash & $0.7931$ & $0.7957$ & $0.7921$ & $0.6018$ & $0.5444$ & $0.7054$\\
North-Small-Translate-1.0 & $0.7952$ & $0.7914$ & $0.7309$ & $0.5272$ & $0.2353$ & $0.6160$\\
Hy-MT2-7B & $0.7334$ & $0.2872$ & $0.0197$ & $0.0006$ & $0.0001$ & $0.2082$\\
Hy-MT2-30B-A3B & $0.7535$ & $0.4566$ & $0.1615$ & $0.0142$ & $0.0664$ & $0.2904$\\
Index-NativeLong-2B & $0.7704$ & $0.7796$ & $0.6508$ & $0.6036$ & $0.5460$ & $0.6701$\\
Index-NativeLong-9B & $0.7836$ & $\mathbf{0.8074}$ & $\mathbf{0.8080}$ & $\mathbf{0.8084}$ & $\mathbf{0.7382}$ & $\mathbf{0.7891}$\\
\midrule
\multicolumn{7}{@{}l}{\textit{BWB (zh$\to$en)}}\\[2pt]
Qwen3.8 Flash & $0.7683$ & $0.7676$ & $0.7387$ & $0.6262$ & $0.5106$ & $0.6823$\\
GLM-5.3-Flash & $0.7632$ & $0.7622$ & $0.7107$ & $0.6344$ & $0.4748$ & $0.6691$\\
North-Small-Translate-1.0 & $0.7717$ & $\mathbf{0.7704}$ & $0.7715$ & $0.6077$ & $0.2196$ & $0.6282$\\
Hy-MT2-7B & $0.7167$ & $0.2387$ & $0.0334$ & $0.0024$ & $0.0001$ & $0.1983$\\
Hy-MT2-30B-A3B & $0.7301$ & $0.4408$ & $0.1432$ & $0.0746$ & $0.1103$ & $0.2998$\\
Index-NativeLong-2B & $0.7688$ & $0.7669$ & $0.7201$ & $0.7100$ & $0.6843$ & $0.7300$\\
Index-NativeLong-9B & $\mathbf{0.7733}$ & $0.7695$ & $\mathbf{0.7790}$ & $\mathbf{0.7756}$ & $\mathbf{0.7440}$ & $\mathbf{0.7683}$\\
\midrule
\multicolumn{7}{@{}l}{\textit{General Books (en$\to$zh)}}\\[2pt]
Qwen3.8 Flash & $0.8945$ & $0.8910$ & $0.8837$ & $0.8582$ & $0.5756$ & $0.8206$\\
GLM-5.3-Flash$^{*}$ & $\mathbf{0.9045}$ & $0.8607$ & $\mathbf{0.8974}$ & $0.8478$ & $\mathbf{0.8871}$ & $0.8795$\\
North-Small-Translate-1.0 & $0.8970$ & $\mathbf{0.8952}$ & $0.8627$ & $0.7873$ & $0.7330$ & $0.8350$\\
Hy-MT2-7B & $0.8246$ & $0.3157$ & $0.0335$ & $0.0004$ & $0.0057$ & $0.2360$\\
Hy-MT2-30B-A3B & $0.8607$ & $0.4395$ & $0.1467$ & $0.0097$ & $0.1962$ & $0.3306$\\
Index-NativeLong-2B & $0.8908$ & $0.8736$ & $0.8735$ & $0.8435$ & $0.8439$ & $0.8651$\\
Index-NativeLong-9B & $0.9016$ & $0.8771$ & $0.8958$ & $\mathbf{0.8980}$ & $0.8514$ & $\mathbf{0.8848}$\\
\bottomrule
\end{tabular}
\caption{Native long-document translation with document-SEGALE/COMET (\texttt{max\_size=8}). Mean equally weights the five Chinese-side length groups. Bold marks the highest displayed score in each column. $^{*}$Some GLM samples were refused; its General Books scores average the remaining samples.}
\label{tab:nailong-frontier}
\end{table}

\FloatBarrier
\Needspace{9\baselineskip}
\section{Analysis}\label{sec:analysis}
\subsection{Mid-training design and language coverage}\label{sec:midtrain-analysis}\label{sec:cpt-recipe}\label{sec:coverage-analysis}
The pilot studies explore data balance and parallel organization using greedy decoding. We use $G$ for general text, $M$ for independently sampled monolingual text, $P$ for ordinary parallel data, and $V$ for pivot-organized data.

For the constant-stage mixture sweep and the stage-progression study, translation scores take the equal-weight average of four dataset/source-language COMET means: FLORES-200 and WMT24++, each with English and Chinese sources. The constant-stage sweep covers 20 FLORES-200 targets and 18 WMT24++ targets. The decay studies use the evaluation scopes specified below and in Appendix~\ref{app:midtrain-pilots}; their translation scores are compared within each study. The 11-task NLU score is the macro-average over MMMLU, Belebele, M3Exam, XCOPA, XCODAH, XCSQA, XStoryCloze, XWinogrande, M\_ARC, MGSM, and INCLUDE, covering multilingual knowledge, reading comprehension, commonsense, and reasoning.

\paragraph{Mixture weights.}
The balanced $G{:}M{:}P=1{:}1{:}1$ constant mixture gives the highest translation score among eight candidates and ties for the highest NLU score. Table~\ref{tab:cpt-mixture-selected} shows four settings, and Appendix~\ref{app:midtrain-pilots} supplies the other four from the same sweep. This pilot supports retaining both monolingual and parallel data. The appendix also reports a separate 4K decay-mixture study, whose translation scores are closely grouped.

\paragraph{Parallel organization.}
The Qwen3-8B parallel/pivot pilot uses a 16K context and compares $G{:}M{:}V=1{:}2{:}4$ with $G{:}M{:}P{:}V=1{:}2{:}2{:}4$ at step 10,000 and the same 25.17B-token decay budget (Table~\ref{tab:cpt-mixture-selected}). Both mixtures include general and monolingual text. The first mixture contains no ordinary parallel data $P$; the second has $P{:}V=1{:}2$ within the full four-stream mixture. On the 20-target-language translation evaluation, the mixture without $P$ scores 0.7991, compared with 0.7971 for the mixture containing $P$. The observed gap of 0.0020 compares the two complete mixtures: including $P$ also changes the normalized sampling shares of $G$, $M$, and $V$. The reported NLU scores are similar. This comparison and the 4K sampling-weight study in Appendix~\ref{app:midtrain-pilots} use different translation evaluation scopes.

\begin{table}[ht]
\centering\small
\begin{tabular}{lrr}
\toprule
Setting & Translation & 11-task NLU\\
\midrule
\multicolumn{3}{l}{\textit{Constant-stage sampling weights ($G{:}M{:}P$), 7k steps}}\\[2pt]
$1{:}1{:}0$ & 0.7750 & 0.7655\\
$1{:}0{:}1$ & 0.7897 & 0.7637\\
$1{:}1{:}1$ & \textbf{0.7914} & \textbf{0.7725}\\
$1{:}1{:}4$ & 0.7907 & 0.7644\\
\midrule
\multicolumn{3}{l}{\textit{Decay-stage sampling weights, 16K context, 25.17B tokens}}\\[2pt]
$G{:}M{:}V=1{:}2{:}4$ & \textbf{0.7991} & 0.7526\\
$G{:}M{:}P{:}V=1{:}2{:}2{:}4$ & 0.7971 & \textbf{0.7554}\\
\bottomrule
\end{tabular}
\caption{Selected mid-training pilots. The constant panel belongs to the eight-setting sweep completed in Appendix~\ref{app:midtrain-pilots}. The parallel/pivot panel uses Qwen3-8B, a 16K context, and a 20-target-language translation evaluation; both runs reach step 10,000 at the same decay budget. Both settings include $G$ and $M$; the first contains no ordinary parallel data $P$. Its translation scores use a different evaluation scope from the appendix decay panel. Bold marks panel maxima.}
\label{tab:cpt-mixture-selected}\label{tab:pivot-organization-matched}
\end{table}

\FloatBarrier
\paragraph{Stage progression.}
Translation improves through both the constant and decay stages at 2B and 9B in the 20+-language development study (Table~\ref{tab:stage-trajectories}). General-task NLU drops during the constant stage and partially recovers during decay, remaining below the base-model score at both sizes.

\begin{table}[ht]
\centering\small
\begin{tabular}{lrrrrrr}
\toprule
 & \multicolumn{3}{c}{Translation} & \multicolumn{3}{c}{11-task NLU}\\
Model & Base & 7k & 10k & Base & 7k & 10k\\
\midrule
Qwen3.5-2B & 0.7015 & 0.7707 & \textbf{0.7898} & \textbf{0.5920} & 0.5303 & 0.5431\\
Qwen3.5-9B & 0.7726 & 0.7924 & \textbf{0.8007} & \textbf{0.7774} & 0.7682 & 0.7712\\
\bottomrule
\end{tabular}
\caption{Training-stage progression in the 20+-language development study, using the metrics defined in Section~\ref{sec:midtrain-analysis}.}
\label{tab:stage-trajectories}
\end{table}

\paragraph{Language expansion.}
Expanding the 2B decay data from 20+ to 100+ languages improves the non-core mean from 0.5605 to 0.6570, with an approximately unchanged core-language mean (Table~\ref{tab:language}). All 61 non-core target languages improve, and their off-target rate falls from 1.94\% to 0.89\%.

\begin{center}
\begin{minipage}{\linewidth}
\centering\small
\begin{tabular}{lrrr}
\toprule
Backbone / mid-training & Core 20 & Non-core 61 & All 81 \\
\midrule
2B base & 0.7319 & 0.5172 & 0.5702 \\
2B, 20+ languages & 0.8231 & 0.5605 & 0.6253 \\
2B, 100+ languages & 0.8219 & \textbf{0.6570} & \textbf{0.6977} \\
9B base & 0.8059 & 0.6952 & 0.7225 \\
9B, 20+ languages & 0.8339 & 0.7141 & 0.7437 \\
\bottomrule
\end{tabular}
\captionof{table}{Language expansion at the mid-training stage on 81 target languages. Corrected COMET averages English- and Chinese-source directions per target, assigning zero to off-target outputs.}
\label{tab:language}
\end{minipage}
\end{center}

\paragraph{Broad language coverage during SFT.}
In a 22-language-only SFT control, 60-language FLORES-200 COMET falls from 0.7441 to 0.7062 between steps 200 and 400, and target-language agreement falls from 59.22\% to 40.02\%, while core-language scores remain nearly unchanged. This failure motivates retaining low-resource directions during SFT and evaluating language coverage beyond the core set.

\FloatBarrier
\Needspace{9\baselineskip}
\subsection{Specialist training and integration}\label{sec:merge-analysis}
\paragraph{SFT-to-RL gains at 9B.}
Table~\ref{tab:9b-sft-rl} compares each specialist before and after RL. For the Instruction expert, instTrans quality improves by 0.0386 and IFscore by 0.0934, with gains on IFMTBench and low-resource instructions. For the General expert, low-resource FLORES COMET-22 improves by 0.0236 and off-target rate falls from 3.4\% to 1.2\%. FLORES XCOMET-XXL and both WMT24++ metrics improve, while FLORES COMET-22 decreases by 0.0017. These results show stronger instruction adherence and low-resource translation after RL, with metric-dependent changes on core-language translation.

\begin{table}[ht]
\centering\small
\begin{tabular}{lrrr}
\toprule
Metric & SFT & RL & $\Delta$ (RL $-$ SFT)\\
\midrule
\multicolumn{4}{l}{\textit{Instruction specialist, 9B}}\\[2pt]
instTrans quality & 0.6320 & 0.6706 & +0.0386\\
instTrans IFscore & 0.7670 & 0.8604 & +0.0934\\
IFMTBench XCOMET-XXL & 0.7804 & 0.7941 & +0.0137\\
IFMTBench IFscore & 0.8662 & 0.8977 & +0.0315\\
Low-resource instTrans quality & 0.4817 & 0.5056 & +0.0239\\
Low-resource instTrans IFscore & 0.7502 & 0.7894 & +0.0392\\
\midrule
\multicolumn{4}{l}{\textit{General specialist, 9B}}\\[2pt]
FLORES COMET-22 & 0.8850 & 0.8833 & -0.0017\\
FLORES XCOMET-XXL & 0.9092 & 0.9140 & +0.0048\\
WMT24++ COMET-22 & 0.8574 & 0.8592 & +0.0018\\
WMT24++ XCOMET-XXL & 0.8840 & 0.8942 & +0.0102\\
Low-resource FLORES COMET-22 & 0.8129 & 0.8365 & +0.0236\\
Low-resource FLORES XCOMET-XXL & 0.6964 & 0.7321 & +0.0357\\
Low-resource off-target rate & 3.4\% & 1.2\% & -2.2 pp\\
\bottomrule
\end{tabular}
\caption{SFT-to-RL changes for the 9B Instruction and General specialists before model integration. Each RL model is compared with its own SFT baseline. Instruction evaluation uses GPT-5.6-Sol. Score differences are absolute; off-target changes are percentage points (pp), where lower is better.}
\label{tab:9b-sft-rl}
\end{table}

\paragraph{Parameter interpolation.}
These ablations use an earlier instruction-expert configuration at the same interpolation weights; the main results use the updated instruction specialist. Instruction scores use GPT-5.6-Sol.
The reference blend improves FLORES and WMT XCOMET-XXL at both sizes, while IFMTBench IFscore declines relative to the general parent; MEME improves at 2B and decreases slightly at 9B (Table~\ref{tab:merge}). A separate 2B, 20-language weight scan supports retaining a dominant general branch: reducing its weight from 1.0 to 0.5 raises WMT XCOMET-XXL from 0.8222 to 0.8522 but lowers IFscore from 0.7395 to 0.2828.

\begin{center}
\begin{minipage}{\linewidth}
\centering\small
\begin{tabular}{llrrrr}
\toprule
Size & Model & FLORES X & WMT X & IFMT IF & MEME \\
\midrule
2B & General & 0.8408 & 0.8336 & \textbf{0.7554} & 0.6034 \\
2B & $0.8G+0.1\,\mathrm{Instruction}+0.1M$ & \textbf{0.8566} & \textbf{0.8419} & 0.7241 & \textbf{0.6429} \\
\midrule
9B & General & 0.8972 & 0.8785 & \textbf{0.8757} & \textbf{0.7383} \\
9B & $0.8G+0.1\,\mathrm{Instruction}+0.1M$ & \textbf{0.9025} & \textbf{0.8819} & 0.8524 & 0.7331 \\
\bottomrule
\end{tabular}
\captionof{table}{Expert integration ablations in the 100+-language setting, using the reference instruction-expert configuration. $G$ denotes General and $M$ denotes Meme. X denotes XCOMET-XXL; IF denotes IFMTBench IFscore. Bold marks the higher value within each model size.}
\label{tab:merge}
\end{minipage}
\end{center}

Syllable control also transfers poorly in a four-way 2B merge: a 0.1-weight video-translation expert yields a SandGlass hit rate of 0.1544, compared with 0.5439 for the specialist.

\paragraph{Targeted MOPD enhancement.}
Larger teachers yield the best instTrans and IFMTBench instruction scores among the three MOPD variants, while all variants reduce low-resource off-target rates from 4.2\% to 2.2\% (Table~\ref{tab:mopd-results}). These gains come with lower translation quality and MEME scores, supporting targeted use for instruction adherence and language consistency.

\begin{center}
\begin{minipage}{\linewidth}
\centering\small
\setlength{\tabcolsep}{4pt}
\begin{tabular}{lrrrrrr}
\toprule
& \multicolumn{2}{c}{instTrans} & IFMTBench & \multicolumn{2}{c}{Low-resource pairs} & MEME \\
\cmidrule(lr){2-3}\cmidrule(lr){5-6}
Model / mixture & Quality $\uparrow$ & IF $\uparrow$ & IF $\uparrow$ & COMET $\uparrow$ & Off-target $\downarrow$ & Score $\uparrow$ \\
\midrule
Reference merge & \textbf{0.5979} & 0.8359 & 0.7241 & 0.7331 & 4.2\% & \textbf{0.6429} \\
\midrule
MOPD (2B, $1{:}1{:}1$) & 0.4848 & 0.8498 & 0.7448 & 0.7511 & \textbf{2.2\%} & 0.6150 \\
MOPD (2B, $1{:}2{:}1$) & 0.5079 & 0.8503 & 0.7470 & \textbf{0.7564} & \textbf{2.2\%} & 0.6068 \\
MOPD (9B$\to$2B, $1{:}1{:}1$) & 0.5572 & \textbf{0.8669} & \textbf{0.7504} & 0.7516 & \textbf{2.2\%} & 0.6275 \\
\bottomrule
\end{tabular}
\captionof{table}{Targeted MOPD for the reference parameter-merge configuration. All evaluated models have 2B parameters. Mixtures are General:Instruction:Meme; 9B$\to$2B denotes the larger-teacher configuration. IF is IFscore. Low-resource pairs are evaluated on FLORES with COMET-22 and off-target rate. Bold marks the best value per column.}
\label{tab:mopd-results}
\end{minipage}
\end{center}

We find that MOPD performs poorly at following length-related constraints. We hypothesize that this limitation may be related to its token-level supervision, and leave further investigation and improved constraint handling to future work.

\FloatBarrier
\Needspace{7\baselineskip}
\subsection{Instruction and length constraints}\label{sec:instruction-analysis}\label{sec:control-generation-analysis}
\paragraph{Constraint composition.}
Index-Translate-9B scores above 2B in all twelve IFMTBench categories, with the largest gaps on constraints combining context and terminology (Table~\ref{tab:ifmt-constraints}). Both models score lower on the combined-constraint subset than on single constraints; all five combined categories include terminology.

\begin{center}
\begin{minipage}{\linewidth}
\centering\small
\begin{tabular}{lrrr}
\toprule
Constraint & Items & 2B & 9B \\
\midrule
\multicolumn{4}{l}{\textit{Single constraints}}\\[2pt]
Code tags & 161 & 0.9938 & 1.0000\\
Inline code & 224 & 0.8929 & 1.0000\\
Layout & 546 & 0.9579 & 0.9689\\
Structured data & 392 & 0.9413 & 0.9974\\
Style & 769 & 0.8252 & 0.8908\\
Terminology & 1,364 & 0.7109 & 0.8364\\
Context & 770 & 0.6522 & 0.8151\\
\addlinespace[2pt]
Weighted mean & 4,226 & 0.7947 & 0.8894\\
\midrule
\multicolumn{4}{l}{\textit{Combined constraints}}\\[2pt]
Structured data + terminology & 567 & 0.8713 & 0.9365\\
Style + terminology & 525 & 0.8179 & 0.9177\\
Context + style + terminology & 619 & 0.6532 & 0.8635\\
Structured data + terminology + style & 637 & 0.6063 & 0.7717\\
Context + terminology & 490 & 0.5808 & 0.7967\\
\addlinespace[2pt]
Weighted mean & 2,838 & 0.7042 & 0.8560\\
\bottomrule
\end{tabular}
\captionof{table}{IFMTBench instruction-following scores for the final Index-Translate models, using GPT-5.6-Sol. Item counts describe the cleaned evaluation subset (7,064 examples); the overall instruction-following score averages 7,063 valid judgments. Section~\ref{sec:eval-setup} explains preprocessing and scoring coverage.}
\label{tab:ifmt-constraints}
\end{minipage}
\end{center}

\paragraph{Context and syllable control.}
On low-resource instTrans, translation quality is 0.3050 for 2B and 0.5222 for 9B, while IFscore is 0.6586 and 0.7725. On FLORES low-resource pairs, XCOMET-XXL is 0.4817 for 2B and 0.6805 for 9B, with off-target rates of 4.2\% and 4.0\%. Index-Translate-35B-A3B (preview) obtains 0.5151 quality and 0.7715 IFscore on low-resource instTrans, with a 4.05\% off-target rate. On FLORES low-resource pairs, it scores 0.8168 COMET-22 and 0.7164 XCOMET-XXL, with a 2.4\% off-target rate. The 9B model improves cross-sentence terminology and coreference resolution in low-resource directions (Table~\ref{tab:instruction-context}). In core directions, subtitle syllable alignment passes in 46.6\% of 2B cases and 64.1\% of 9B cases, while the other hard-constraint pass rates exceed 93\%. This source-relative subtitle constraint complements the explicit target-count control studied with Index-Homura.

\begin{center}
\begin{minipage}{\linewidth}
\centering\small
\setlength{\tabcolsep}{4pt}
\begin{tabular}{lrrrrr}
\toprule
& & \multicolumn{2}{c}{Core directions} & \multicolumn{2}{c}{Low-resource directions} \\
\cmidrule(lr){3-4}\cmidrule(lr){5-6}
Constraint & Items (core / low) & 2B & 9B & 2B & 9B \\
\midrule
Subtitle syllable alignment & 601 / 765 & 46.6\% & 64.1\% & 72.8\% & 84.6\%\\
Style consistency & 633 / 988 & 0.7157 & 0.8525 & 0.5220 & 0.7364\\
Cross-sentence terminology & 508 / 777 & 0.6486 & 0.7139 & 0.4027 & 0.6536\\
Coreference resolution & 24 / 546 & 0.6667 & 0.6250 & 0.5250 & 0.7542\\
Context disambiguation & 42 / 54 & 0.6026 & 0.7439 & 0.5566 & 0.7593\\
\bottomrule
\end{tabular}
\captionof{table}{Selected constraints in instTrans and its low-resource extension. Syllable alignment is a binary-check pass rate; the remaining rows are mean soft-constraint scores from GPT-5.6-Sol. Item counts are per constraint, and a sample can contain multiple constraints.}
\label{tab:instruction-context}
\end{minipage}
\end{center}

\paragraph{Quality--control trade-off.}
Increasing the 2B syllable-reward weight from 1 to 2 raises the SandGlass within-10\% rate from 63.08\% to 74.19\%, while translation quality falls from 0.7615 to 0.6900. Appendix~\ref{app:translation-cases} provides representative instruction-translation cases.

\begin{figure}[!htbp]
\centering
\includegraphics[width=0.96\linewidth]{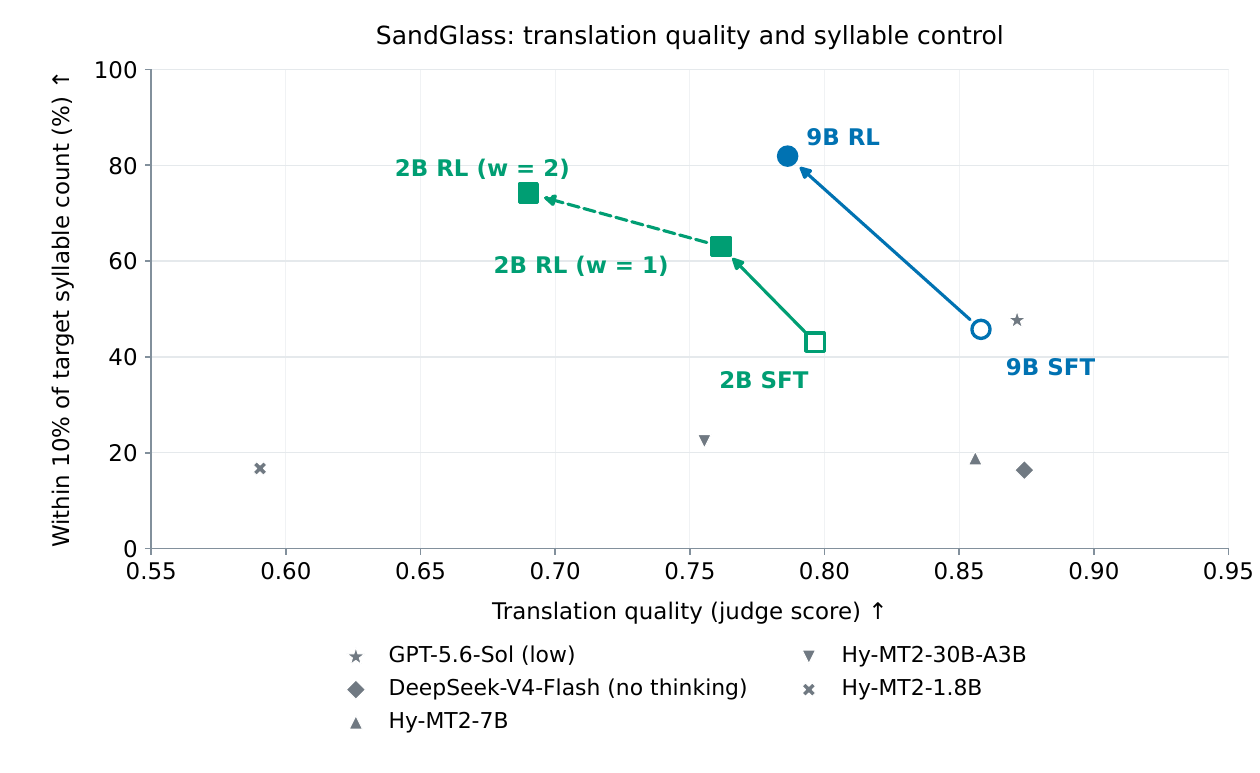}
\caption{Quality and syllable-control trade-offs on SandGlass. Points reproduce Index-Homura and selected baselines from Table~\ref{tab:homura-expanded}; the additional 2B point with syllable-reward weight $w=2$ comes from the study in this section. Blue and green denote the 9B and 2B Homura models. Solid arrows connect SFT to RL, and the dashed arrow connects the two 2B reward weights. Both axes favor larger values; arrows show changes between evaluated configurations.}
\label{fig:homura-quality-control}
\end{figure}

\FloatBarrier
\Needspace{6\baselineskip}
\noindent\begin{minipage}{\linewidth}
\section{Conclusion}
Index-Translate is a multilingual translation model family that combines multilingual mid-training and specialist integration for general translation and instruction following across 150 languages. Our experiments highlight the importance of broad language coverage and the trade-offs between translation quality and constraint adherence. Text-only mid- and post-training also support image-text translation through the visual pathway inherited from Qwen3.5 base models, with stronger cross-modal generalization at larger scales. Index-Echo, Index-Homura, and Index-NativeLong extend the shared foundation through dedicated training for speech, syllable control, and native long-document translation.
\end{minipage}

\clearpage
\begingroup\footnotesize\raggedright

\endgroup

\clearpage
\appendix
\section{Supplementary Training Studies}
\subsection{Parallel and pivot organization}\label{app:parallel-schematic}
Let $x_\ell$ denote the same content expressed in language $\ell$. Ordinary parallel data align a language pair directly. Pivot organization connects multiple translations through a shared English anchor, as illustrated in Table~\ref{tab:parallel-schematic}. Parallel degree counts aligned language variants, whereas $P{:}V$ specifies the sampling ratio between the two data streams.

\begin{table}[ht]
\centering\small
\begin{tabularx}{\linewidth}{@{}l>{\raggedright\arraybackslash}Xl@{}}
\toprule
Organization & Illustrative alignment & Parallel degree\\
\midrule
Parallel ($P$) & $x_{\mathrm{zh}}\longleftrightarrow x_{\mathrm{ja}}$ & 2 languages\\[4pt]
Pivot ($V$) & $x_{\mathrm{zh}}\longleftrightarrow x_{\mathrm{en}}\longleftrightarrow x_{\mathrm{ja}}$ & 3 languages\\
\bottomrule
\end{tabularx}
\caption{Illustrative parallel and pivot groups. Arrows denote cross-language alignment; the pivot example shares one English anchor across the Chinese and Japanese variants.}
\label{tab:parallel-schematic}
\end{table}

\subsection{Additional mid-training mixtures}\label{app:midtrain-pilots}
The constant-stage rows in Table~\ref{tab:cpt-mixture} complete the same eight-setting $G{:}M{:}P$ sweep at step 7,000 as the upper panel of Table~\ref{tab:cpt-mixture-selected}. $G$, $M$, $P$, and $V$ denote general, monolingual, parallel, and pivot data.

The decay rows are a separate 4K sampling-weight study at step 10,000, varying how pivot data are mixed with general and monolingual text. Their translation score takes the equal-weight average of six dataset/source-language COMET means: WMT24++, FLORES-200, and BOUQuET, each with English and Chinese sources and eight target languages. The $G{:}M{:}V=1{:}2{:}4$ row uses the same sampling weights as the setting without ordinary parallel data in Table~\ref{tab:cpt-mixture-selected}, but the latter uses a 16K context and reports translation over 20 target languages. Translation scores should therefore be compared within each study. Each panel compares configurations at the same training step; stage-to-stage trajectories are reported separately in Table~\ref{tab:stage-trajectories}.

\begin{table}[ht]
\centering\small
\begin{tabular}{llrr}
\toprule
Phase & Sampling weights & Translation & 11-task NLU\\
\midrule
Constant, 4K, 7k & $G{:}M{:}P=1{:}1{:}2$ & 0.7903 & 0.7638\\
 & $G{:}M{:}P=1{:}2{:}1$ & 0.7898 & 0.7711\\
 & $G{:}M{:}P=2{:}1{:}1$ & 0.7873 & 0.7641\\
 & $G{:}M{:}P=2{:}2{:}1$ & 0.7879 & 0.7725\\
\midrule
Decay, 4K, 10k & $G{:}M{:}V=1{:}2{:}2$ & 0.8023 & 0.7763\\
 & $G{:}M{:}V=1{:}2{:}4$ & 0.8033 & 0.7717\\
 & $G{:}V=1{:}8$ & 0.8019 & 0.7702\\
 & $G{:}V=1{:}4$ & 0.8027 & 0.7710\\
\bottomrule
\end{tabular}
\caption{Additional pilot mixtures. Constant rows complete the sweep in Table~\ref{tab:cpt-mixture-selected}. Decay rows form a separate 4K study using the eight-target-language, six-component translation mean defined in Appendix~\ref{app:midtrain-pilots}. NLU is the 11-task macro-average defined in Section~\ref{sec:midtrain-analysis}. Compare translation scores within each study.}
\label{tab:cpt-mixture}
\end{table}

\clearpage
\subsection{NativeLong training ablations}\label{app:nailong-ablation}
We study decay sequence length, position encoding, and length generalization in separate 2B experiments using GuoFeng-only document SFT (Table~\ref{tab:nailong-ablation}). Scores use the same 0--1 scale as the main evaluation; each study averages over its own evaluation passages.

\paragraph{Long-sequence decay.}
With matched short document SFT, using prepacked 128K rather than 4K sequences during mid-training decay improves mean quality by 0.0528 with RoPE~\cite{su2021roformer} and 0.0495 with NoPE (no explicit positional encoding)~\cite{kazemnejad2023nope}. The 128K route uses exponential learning-rate decay under WSD.

\paragraph{Position encoding and length generalization.}
After document SFT covering source lengths up to approximately 64K, the two 128K-decay bases have similar overall scores: approximately 0.7810 for RoPE and 0.7819 for NoPE. Their strengths vary by length, with NoPE ahead at 16K and RoPE ahead at 64K. When document SFT is limited to source texts of approximately 16K or less, NoPE scores higher over the 4K--32K evaluation range, with a 0.0370 advantage at 32K.

\begin{table}[!ht]
\centering\footnotesize
\setlength{\tabcolsep}{7pt}
\renewcommand{\arraystretch}{1.1}
\begin{tabular}{@{}lrrr@{}}
\toprule
\multicolumn{4}{@{}l}{\textit{(a) Decay sequence length, followed by matched short document SFT}}\\[2pt]
Position encoding & 4K decay & 128K decay & Gain\\
\midrule
RoPE & 0.6673 & \textbf{0.7201} & +0.0528\\
NoPE & 0.6739 & \textbf{0.7234} & +0.0495\\
\bottomrule
\end{tabular}

\medskip
\begin{tabular}{@{}lrrrr@{}}
\toprule
& \multicolumn{2}{c}{\textit{(b) Document SFT up to 64K}} & \multicolumn{2}{c}{\textit{(c) Document SFT up to 16K}}\\
\cmidrule(lr){2-3}\cmidrule(l){4-5}
Test length & RoPE & NoPE & RoPE & NoPE\\
\midrule
4K & 0.7930 & \textbf{0.7950} & 0.7910 & \textbf{0.7918}\\
8K & \textbf{0.8000} & 0.7970 & 0.7839 & \textbf{0.7974}\\
16K & 0.6780 & \textbf{0.7910} & 0.7318 & \textbf{0.7586}\\
32K & \textbf{0.8050} & 0.7920 & 0.6264 & \textbf{0.6634}\\
64K & \textbf{0.7670} & 0.6780 & -- & --\\
\midrule
Overall & $\approx0.7810$ & $\mathbf{\approx0.7819}$ & 0.7513 & \textbf{0.7667}\\
\bottomrule
\end{tabular}
\caption{GuoFeng 2B ablations using document-SEGALE/COMET (\texttt{max\_size=8}) on a 0--1 scale. Panels (b) and (c) start from 128K-decay bases. Each overall score averages the evaluation passages in its own study.}
\label{tab:nailong-ablation}
\end{table}

\clearpage
\section{Supplementary Evaluation Details and Cases}\label{app:translation-cases}
\subsection{Translation with task constraints}
We present two examples from the instruction-following evaluations, showing the source, task requirements, and recorded Index-Translate-9B output.

\paragraph{Cross-sentence terminology.}
The Chinese-to-Russian subtitle task asks the model to use consistent translations for cracks, carbon fiber, and concrete. Table~\ref{tab:translation-terminology-case} shows three segments in which the translation preserves the technical vocabulary, with grammatical inflection across sentences.

\begin{center}
\begin{minipage}{\linewidth}
\small
\begin{tabularx}{\linewidth}{@{}l>{\raggedright\arraybackslash}X@{}}
\toprule
Segment & Source and Index-Translate-9B translation \\
\midrule
2 & \zh{王平仲却采用了，更加昂贵的碳纤维材料}\\[2pt]
 & Ван Пинчжун решил использовать более дорогой материал — углеродное волокно.\\[5pt]
5 & \zh{碳纤维的好处呢，就是它抗裂缝}\\[2pt]
 & Преимущество углеродного волокна в том, что оно противостоит трещинам.\\[5pt]
6 & \zh{而且混凝土的刚性，老的混凝土以前不是裂缝嘛}\\[2pt]
 & К тому же, что касается жёсткости бетона — ведь в старом бетоне раньше были трещины, не так ли?\\[5pt]
\bottomrule
\end{tabularx}
\captionof{table}{Chinese-to-Russian subtitle excerpts from instTrans. Three segments from the same ten-segment input illustrate terminology consistency; source text and translations are reproduced verbatim.}
\label{tab:translation-terminology-case}
\end{minipage}
\end{center}

\paragraph{Structured-data translation.}
The English-to-Chinese task asks for translation of the Category, Description, and Release\_Status columns, preserving the CSV header, field boundaries, product IDs, and prices. Table~\ref{tab:translation-csv-case} reproduces the complete input and output.

\begin{center}
\begin{minipage}{\linewidth}
\small
\textbf{Source (English)}\par\smallskip
{\ttfamily\footnotesize\raggedright
Product\_ID,Category,Description,Release\_Status,Price\par\smallskip
"SYS-001","Hardware","High-performance motherboard, equipped with dual Gigabit LAN and PCIe 4.0 slots.","Available",199.99\par\smallskip
"SYS-002","Software","Enterprise-level firewall solution, providing real-time threat detection and automated response protocols.","In Beta",49.99\par\smallskip
}\medskip
\textbf{Index-Translate-9B (Chinese)}\par\smallskip
{\chinesefont\small\raggedright\XeTeXlinebreaklocale "zh"\XeTeXlinebreakskip=0pt plus 1pt
Product\_ID,Category,Description,Release\_Status,Price\par\smallskip
"SYS-001","硬件","高性能主板，配备双千兆局域网和PCIe 4.0插槽。","已发布",199.99\par\smallskip
"SYS-002","软件","企业级防火墙解决方案，提供实时威胁检测和自动化响应协议。","测试版",49.99\par\smallskip
}
\captionof{table}{English-to-Chinese structured translation from IFMTBench. The header, product IDs, prices, and CSV field boundaries are preserved while the three requested text columns are translated.}
\label{tab:translation-csv-case}
\end{minipage}
\end{center}

\clearpage
\subsection{Meme translation cases}\label{app:meme-cases}
Table~\ref{tab:meme-cases} illustrates nonstandard spelling, a community-specific alias, and a laughter expression. Both Index-Translate sizes recover the intended meanings in these selected examples.

\begin{center}
\begin{minipage}{\linewidth}
\centering\small
\setlength{\tabcolsep}{5pt}
\renewcommand{\arraystretch}{1.08}
\begin{tabularx}{\linewidth}{@{}l>{\raggedright\arraybackslash}X@{}}
\toprule
Model & Recorded translation \\
\midrule
\multicolumn{2}{@{}p{\linewidth}@{}}{\textbf{1. Playful proposal.} Source: \zh{那个，你好，结芬}\texttt{<EMOTE-placeholder-0>}}\\[2pt]
\multicolumn{2}{@{}p{\linewidth}@{}}{\textit{Intended meaning:} A playful request to get married, rather than a personal name.}\\[3pt]
GPT-5.6-Sol & Um, hello, Jie Fen \texttt{<EMOTE-placeholder-0>}\\
Hy-MT2-7B & Hey there, Jiefen\texttt{<EMOTE-placeholder-0>}\\
Index-Translate-2B & Um, hi there, let's get married\texttt{<EMOTE-placeholder-0>}\\
Index-Translate-9B & Um, hi, let's get married \texttt{<EMOTE-placeholder-0>}\\
\midrule
\multicolumn{2}{@{}p{\linewidth}@{}}{\textbf{2. Gaming-community alias.} Source: \zh{狒瘾犯了就去打}}\\[2pt]
\multicolumn{2}{@{}p{\linewidth}@{}}{\textit{Intended meaning:} An urge to play Final Fantasy XIV (FFXIV).}\\[3pt]
GPT-5.6-Sol & If you’re itching for more JoJo, go play the game.\\
Hy-MT2-7B & If you get monkey addiction, go fight.\\
Index-Translate-2B & Go play when my FFXIV addiction kicks in.\\
Index-Translate-9B & When the FFXIV itch hits, just go play.\\
\midrule
\multicolumn{2}{@{}p{\linewidth}@{}}{\textbf{3. Laughter expression.} Source: Fuyuan Flower\zh{红红火火恍恍惚惚}}\\[2pt]
\multicolumn{2}{@{}p{\linewidth}@{}}{\textit{Intended meaning:} The repeated pinyin initial h conveys laughter, like repeated ``ha''.}\\[3pt]
GPT-5.6-Sol & Fuyuan Flower—red-hot and thriving, hazy and wavy.\\
Hy-MT2-7B & Fuyuan Flower: booming yet confused.\\
Index-Translate-2B & Fuyuan Flower, hahaha, what a wild ride!\\
Index-Translate-9B & Fuyuan Flower: Hahahahaha, this is wild!\\
\bottomrule
\end{tabularx}
\captionof{table}{Selected MEME examples from the final Index-Translate models. Both Index-Translate sizes receive a judge score of 1 on each example; the displayed baselines receive 0. Intended meanings summarize benchmark annotations. Translations and placeholders are reproduced from the recorded outputs.}
\label{tab:meme-cases}
\end{minipage}
\end{center}

\clearpage
\subsection{Syllable-controlled translation cases}\label{app:homura-cases}
Table~\ref{tab:homura-cases} shows Index-Homura-9B translating the same source at three target syllable counts. The English example varies introductory wording and noun phrases; the Japanese example uses increasingly expanded expressions for a deeply ingrained traumatic memory.

\begin{center}
\begin{minipage}{\linewidth}
\centering\small
\setlength{\tabcolsep}{5pt}
\renewcommand{\arraystretch}{1.18}
\begin{tabularx}{\linewidth}{@{}lrr>{\raggedright\arraybackslash}X@{}}
\toprule
Budget & Target & Actual & Recorded translation\\
\midrule
\multicolumn{4}{@{}p{\linewidth}@{}}{\textbf{Chinese to English.} Source: \zh{说到底聊天群的规则一句话就能总结}}\\[4pt]
Short & 10 & 10 & The group's rules can be summed up in one line.\\
Natural & 14 & 14 & Ultimately, the group's rules can be summed up in one line.\\
Long & 18 & 18 & Ultimately, the rules of the chat group can be summed up in one sentence.\\
\midrule
\multicolumn{4}{@{}p{\linewidth}@{}}{\textbf{Chinese to Japanese.} Source: \zh{先前被虐的阴影可谓是印刻到了骨子里头了}}\\[4pt]
Short & 15 & 15 & \ja{虐められたトラウマが根深い。}\\
Natural & 20 & 20 & \ja{前のトラウマが骨に刻まれてるんだな}\\
Long & 25 & 25 & \ja{前の屈辱が骨の髄まで刻み込まれていたな。}\\
\bottomrule
\end{tabularx}
\captionof{table}{Selected Index-Homura-9B examples from SandGlass. Target and actual syllable counts follow the benchmark's counting procedure. Each displayed output exactly meets its target and receives a Gemini-2.5-Flash quality score of 1.0. Translations are reproduced verbatim.}
\label{tab:homura-cases}
\end{minipage}
\end{center}

\clearpage
\subsection{Long-document translation cases}\label{app:longdoc-cases}
\paragraph{Comparison setup.}
The three documents in Section~\ref{sec:nailong-training} are held out by work. We compare native Index-NativeLong-9B with three TranslateBooksWithLLMs configurations (version 1.5.13): the same 9B model with an automatic glossary, Qwen3.8-Flash with an automatic glossary, and Flash with a glossary generated by GPT-6-Sol from the full source. Automatic glossaries are generated by the translating model from source excerpts sampled by the framework. All glossaries use source text alone. Chunked translation receives neighboring source text and the preceding translation's tail.

\paragraph{Glacial history: preserving the author's concepts.}
This English-to-Chinese example comes from the 32K group of General Books. Around 10.4\% into the source, the author defines an \emph{ice age} as a whole cold episode that includes both \emph{glacials} and \emph{interglacials}. Table~\ref{tab:glacial-case} presents the source and four translations of this passage. The chunked 9B run uses temperature 0.2 and repetition penalty 1.05.

\begin{center}
\begin{minipage}{\linewidth}
\centering\small
\setlength{\tabcolsep}{4pt}
\renewcommand{\arraystretch}{1.18}
\begin{tabularx}{\linewidth}{@{}>{\raggedright\arraybackslash}p{0.22\linewidth} >{\raggedright\arraybackslash}X@{}}
\toprule
Configuration & Source or translation excerpt\\
\midrule
Source & In current usage, the term ``ice age'' properly refers to an entire cold episode, including its short warm periods. [\ldots] Large-scale advances and retreats of ice during an ice age are usually referred to as glacials and interglacials respectively.\\[4pt]
Native 9B & \zh{按当前用法，“冰河时代”一词应指整个寒冷事件，包括其中的短暂温暖期。}[\ldots]\zh{冰河时代中冰层的大规模进退，通常分别称为冰期和间冰期。}\\[4pt]
Chunked 9B + automatic glossary & \zh{按照当前用法，“冰河时代”一词应指整个寒冷时期，包括其中短暂的温暖期。}[\ldots]\zh{冰河时代中冰层的大规模推进和退缩，通常分别称为冰期和间冰期。}\\[4pt]
Chunked Flash + automatic glossary & \zh{在当今的用法中，“冰期”一词恰当地指代整个寒冷时期，包括其间短暂的温暖阶段。}[\ldots]\zh{在一个冰期内，大规模冰川推进和消退通常分别被称为冰期和间冰期。}\\[4pt]
Chunked Flash + full-source glossary & \zh{在当前的用法中，“冰期”一词确切地指代整个寒冷时期，包括其间短暂的温暖阶段。}[\ldots]\zh{在冰期内，大规模冰川的推进和消退通常分别被称为冰期和间冰期。}\\
\bottomrule
\end{tabularx}
\captionof{table}{Glacial-history excerpts at the same source passage. Bracketed ellipses omit intervening text. Both 9B translations preserve the whole--part distinction; both Flash translations use \zh{冰期} for the whole episode and its cold stages.}
\label{tab:glacial-case}
\end{minipage}
\end{center}

The glossary conditions differ. The 9B automatic glossary contains \emph{Ice Age} $\to$ Ice Age, leaving the term in English; Flash's automatic glossary omits \emph{ice age}, \emph{glacial}, and \emph{interglacial}. The full-source glossary supplies \emph{ice age} $\to$ \zh{冰期}, \emph{glacial} $\to$ \zh{冰期（冰进期）}, and \emph{interglacial} $\to$ \zh{间冰期}. Thus, the two Flash outputs lose the distinction under different glossary conditions: a fixed term list, even one prepared from the full source, does not ensure that the translation preserves the author's conceptual hierarchy.

\paragraph{Entity consistency.}
Table~\ref{tab:longdoc-cases} pairs source excerpts with recurring names in the four translations. The early mention identifies \zh{王妃} as a mentor. Her name is missing from the 9B glossary, mapped to Princess Consort in Flash's automatic glossary, and correctly supplied as Wang Fei by the full-source glossary. All three chunked glossaries omit Zeitlin.

\begin{center}
\begin{minipage}{\linewidth}
\centering\small
\setlength{\tabcolsep}{4pt}
\renewcommand{\arraystretch}{1.18}
\begin{tabularx}{\linewidth}{@{}>{\raggedright\arraybackslash}p{0.30\linewidth} >{\raggedright\arraybackslash}X >{\raggedright\arraybackslash}X >{\raggedright\arraybackslash}X@{}}
\toprule
Configuration & Early mention & Middle mention & Late mention\\
\midrule
\multicolumn{4}{@{}p{\linewidth}@{}}{\textbf{Character identity: \zh{王妃}.} Chinese to English; positions 31.3\%, 56.6\%, and 84.8\%.}\\[4pt]
Source excerpt & \zh{王妃导师让我来找你过去一趟} & \zh{又被王妃叫去了办公室} & \zh{王妃看到圣枪骑士的时候}\\[3pt]
Full-document 9B & Mentor Wang Fei & Wang Fei & Wang Fei\\
Chunked 9B + automatic glossary & Instructor Wangfei & the Dean & The Queen Consort\\
Chunked Flash + automatic glossary & Princess Consort & the Princess Consort & Princess Consort\\
Chunked Flash + full-source glossary & Instructor Wang Fei & Wang Fei & Wang Fei\\
\midrule
\multicolumn{4}{@{}p{\linewidth}@{}}{\textbf{Name consistency: Zeitlin.} English to Chinese; positions 32.9\%, 38.7\%, and 68.2\%.}\\[4pt]
Source excerpt & a perspective Zeitlin has termed & Zeitlin and Tolli-day\textquotesingle s work & As Zeitlin has reminded us\\[3pt]
Full-document 9B & \zh{泽特林} & \zh{泽特林} & \zh{泽特林}\\
Chunked 9B + automatic glossary & \zh{泽特林} & Zeitlin & \zh{泽特林}\\
Chunked Flash + automatic glossary & \zh{齐特林（}Zeitlin\zh{）} & \zh{齐特林} & \zh{蔡特林（}Zeitlin\zh{）}\\
Chunked Flash + full-source glossary & \zh{齐特林} & \zh{齐特林} & \zh{蔡特林（}Zeitlin\zh{）}\\
\bottomrule
\end{tabularx}
\captionof{table}{Selected entity mentions under four long-document translation configurations. Positions are measured by source-character offset; excerpts omit surrounding text. Dean and Queen Consort change the character\textquotesingle s identity, while Wangfei and Wang Fei differ only in spacing.}
\label{tab:longdoc-cases}
\end{minipage}
\end{center}

\paragraph{Translation workflow.}
Across the three documents, native 9B uses three model calls and 90,143 input tokens, compared with 160 calls and 225,125 input tokens for the same model using automatic glossaries and chunked translation. The difference reflects extra calls and repeated context in the chunked workflow.

\noindent\begin{minipage}{\linewidth}
\subsection{Judge failure cases}\label{app:judge-cases}\label{sec:reward-hacking-analysis}
Policy outputs sometimes append unsupported markers such as \texttt{[1]} and \texttt{/ End/}, without adding source-grounded meaning (Table~\ref{tab:judge-failures}). Reward optimization can amplify evaluator errors, and adversarial suffixes can inflate LLM-judge scores~\cite{gao2023reward,raina2024judge}. These observations motivate the adversarial judge supervision in Section~\ref{sec:robust-judge}.

\begin{center}
\centering\small
\setlength{\tabcolsep}{5pt}
\renewcommand{\arraystretch}{1.15}
\begin{tabular}{@{}p{0.19\linewidth}p{0.32\linewidth}p{0.42\linewidth}@{}}
\toprule
Pattern & Source meaning & Recorded output\\
\midrule
Citation suffix & It probably will not be a big event. & No será un gran acontecimiento. \texttt{[1]}\\
Termination marker & Pulled him out of that rubbish. & Lo saqué de entre toda esa porquería. \texttt{/ End/}\\
\bottomrule
\end{tabular}
\captionof{table}{Representative unsupported suffixes in policy outputs. Source meanings are English glosses of the Chinese inputs; Spanish outputs preserve the recorded markers.}
\label{tab:judge-failures}
\end{center}

\end{minipage}

\Needspace{8\baselineskip}
\subsection{IFMTBench preprocessing and scoring}\label{app:ifmt-preprocessing}
The original IFMTBench contains 7,344 examples: 4,506 single-constraint and 2,838 combined-constraint items~\cite{zheng2026hymt2}. Our evaluation uses a cleaned subset of 7,064 examples (4,226 single-constraint and 2,838 combined-constraint items), after removing 280 single-constraint examples with data-quality issues. These include corrupted reference text, where unassigned or private-use code points replace or interrupt ordinary words, and inconsistencies between language labels and the source or reference language. The reported instruction-following aggregate averages 7,063 valid judgments; one retained example did not yield a valid judgment.

For combined constraints, hard checks gate the mean graded constraint score. IFscore is the sample-level mean over valid judgments, and reference-based XCOMET-XXL measures translation quality.

\clearpage
\section{Supported Languages and Identifiers}\label{app:language-inventory}
The supported-language inventory contains 150 entries, including English and Chinese: 22 core and 128 additional entries. Regional and script-specific varieties are counted separately as listed. Four pairs of alternate codes each identify one entry, yielding 154 identifiers in total. Tables~\ref{tab:language-inventory-core} and~\ref{tab:language-inventory-additional} provide the complete inventory with these alternate codes grouped together. Benchmark subsets are specified separately in Section~\ref{sec:eval-setup}.

\begingroup
\fontsize{8.5}{10.5}\selectfont
\setlength{\tabcolsep}{3pt}
\renewcommand{\arraystretch}{1.10}
\setlength{\LTleft}{0pt}
\setlength{\LTright}{0pt plus 1fill}
\setlength{\LTcapwidth}{\textwidth}
\begin{longtable}{@{}>{\raggedright\arraybackslash}p{0.235\textwidth}>{\raggedright\arraybackslash}p{0.24\textwidth}>{\raggedright\arraybackslash}p{0.235\textwidth}>{\raggedright\arraybackslash}p{0.24\textwidth}@{}}
\caption{Core languages: 22 entries, including English and Chinese. Alternate identifiers are grouped together.}\label{tab:language-inventory-core}\\
\toprule
Code(s) & Language / variety & Code(s) & Language / variety \\
\midrule
\endfirsthead
\multicolumn{4}{@{}l}{\textit{Table \thetable\ continued}}\\
\toprule
Code(s) & Language / variety & Code(s) & Language / variety \\
\midrule
\endhead
\midrule
\multicolumn{4}{r@{}}{\textit{Continued on next page}}\\
\endfoot
\bottomrule
\endlastfoot
\texttt{en}, \texttt{eng\_Latn} & English & \texttt{zh}, \texttt{zho\_Hans} & Chinese \\
\texttt{de} & German & \texttt{fr} & French \\
\texttt{es} & Spanish & \texttt{ja} & Japanese \\
\texttt{ko} & Korean & \texttt{pt} & Portuguese \\
\texttt{ru} & Russian & \texttt{ar} & Arabic \\
\texttt{it} & Italian & \texttt{nl} & Dutch \\
\texttt{pl} & Polish & \texttt{ro} & Romanian \\
\texttt{sv} & Swedish & \texttt{tr} & Turkish \\
\texttt{hi} & Hindi & \texttt{vi} & Vietnamese \\
\texttt{th} & Thai & \texttt{id} & Indonesian \\
\texttt{ms} & Malay & \texttt{fil} & Filipino \\
\end{longtable}
\endgroup

\begingroup
\fontsize{8.5}{10.5}\selectfont
\setlength{\tabcolsep}{3pt}
\renewcommand{\arraystretch}{1.10}
\setlength{\LTleft}{0pt}
\setlength{\LTright}{0pt plus 1fill}
\setlength{\LTcapwidth}{\textwidth}
\begin{longtable}{@{}>{\raggedright\arraybackslash}p{0.235\textwidth}>{\raggedright\arraybackslash}p{0.24\textwidth}>{\raggedright\arraybackslash}p{0.235\textwidth}>{\raggedright\arraybackslash}p{0.24\textwidth}@{}}
\caption{Additional languages and varieties: 128 entries. Alternate identifiers are grouped together.}\label{tab:language-inventory-additional}\\
\toprule
Code(s) & Language / variety & Code(s) & Language / variety \\
\midrule
\endfirsthead
\multicolumn{4}{@{}l}{\textit{Table \thetable\ continued}}\\
\toprule
Code(s) & Language / variety & Code(s) & Language / variety \\
\midrule
\endhead
\midrule
\multicolumn{4}{r@{}}{\textit{Continued on next page}}\\
\endfoot
\bottomrule
\endlastfoot
\texttt{ukr\_Cyrl} & Ukrainian & \texttt{fas\_Arab}, \texttt{pes\_Arab} & Persian \\
\texttt{ces\_Latn} & Czech & \texttt{ell\_Grek} & Greek \\
\texttt{dan\_Latn} & Danish & \texttt{hun\_Latn} & Hungarian \\
\texttt{fin\_Latn} & Finnish & \texttt{nob\_Latn} & Norwegian Bokmål \\
\texttt{slk\_Latn} & Slovak & \texttt{bul\_Cyrl} & Bulgarian \\
\texttt{bos\_Latn} & Bosnian & \texttt{cat\_Latn} & Catalan \\
\texttt{ben\_Beng} & Bengali & \texttt{heb\_Hebr} & Hebrew \\
\texttt{lit\_Latn} & Lithuanian & \texttt{slv\_Latn} & Slovenian \\
\texttt{ekk\_Latn} & Estonian & \texttt{als\_Latn} & Albanian \\
\texttt{lvs\_Latn} & Latvian & \texttt{azj\_Latn} & Azerbaijani \\
\texttt{hrv\_Latn} & Croatian & \texttt{tam\_Taml} & Tamil \\
\texttt{npi\_Deva} & Nepali & \texttt{urd\_Arab} & Urdu (Arabic script) \\
\texttt{mkd\_Cyrl} & Macedonian & \texttt{srp\_Cyrl} & Serbian (Cyrillic) \\
\texttt{mar\_Deva} & Marathi & \texttt{kat\_Geor} & Georgian \\
\texttt{mal\_Mlym} & Malayalam & \texttt{kaz\_Cyrl} & Kazakh \\
\texttt{isl\_Latn} & Icelandic & \texttt{glg\_Latn} & Galician \\
\texttt{kan\_Knda} & Kannada & \texttt{ary\_Arab} & Moroccan Arabic \\
\texttt{guj\_Gujr} & Gujarati & \texttt{bel\_Cyrl} & Belarusian \\
\texttt{afr\_Latn} & Afrikaans & \texttt{hye\_Armn} & Armenian \\
\texttt{khk\_Cyrl} & Mongolian & \texttt{khm\_Khmr} & Khmer \\
\texttt{eus\_Latn} & Basque & \texttt{mya\_Mymr} & Burmese \\
\texttt{lat\_Latn} & Latin & \texttt{uzn\_Cyrl} & Uzbek (Cyrillic) \\
\texttt{tel\_Telu} & Telugu & \texttt{ory\_Orya} & Odia \\
\texttt{nno\_Latn} & Norwegian Nynorsk & \texttt{uzn\_Latn} & Uzbek (Latin) \\
\texttt{swh\_Latn} & Swahili & \texttt{sin\_Sinh} & Sinhala \\
\texttt{kir\_Cyrl} & Kyrgyz & \texttt{som\_Latn} & Somali \\
\texttt{pan\_Guru} & Punjabi & \texttt{arz\_Arab} & Egyptian Arabic \\
\texttt{cym\_Latn} & Welsh & \texttt{nrm\_Latn} & Norman \\
\texttt{pbt\_Arab} & Pashto & \texttt{gle\_Latn} & Irish \\
\texttt{srp\_Latn} & Serbian (Latin) & \texttt{hau\_Latn} & Hausa \\
\texttt{ckb\_Arab} & Central Kurdish & \texttt{mlt\_Latn} & Maltese \\
\texttt{yue\_Hani}, \texttt{yue\_Hant} & Cantonese & \texttt{tgk\_Cyrl} & Tajik \\
\texttt{kmr\_Latn} & Northern Kurdish & \texttt{bew\_Latn} & Betawi \\
\texttt{amh\_Ethi} & Amharic & \texttt{lao\_Laoo} & Lao \\
\texttt{ltz\_Latn} & Luxembourgish & \texttt{fry\_Latn} & Frisian \\
\texttt{div\_Thaa} & Dhivehi & \texttt{epo\_Latn} & Esperanto \\
\texttt{kin\_Latn} & Kinyarwanda & \texttt{ars\_Arab} & Najdi Arabic \\
\texttt{fao\_Latn} & Faroese & \texttt{plt\_Latn} & Malagasy \\
\texttt{asm\_Beng} & Assamese & \texttt{snd\_Arab} & Sindhi \\
\texttt{xho\_Latn} & Xhosa & \texttt{tuk\_Latn} & Turkmen \\
\texttt{hat\_Latn} & Haitian Creole & \texttt{gla\_Latn} & Scottish Gaelic \\
\texttt{ceb\_Latn} & Cebuano & \texttt{ydd\_Hebr} & Yiddish \\
\texttt{jav\_Latn} & Javanese & \texttt{bak\_Cyrl} & Bashkir \\
\texttt{pap\_Latn} & Papiamento & \texttt{mri\_Latn} & Māori \\
\texttt{uig\_Arab} & Uyghur & \texttt{tat\_Cyrl} & Tatar \\
\texttt{bod\_Tibt} & Tibetan & \texttt{mww\_Latn} & White Hmong \\
\texttt{hyw\_Armn} & Western Armenian & \texttt{yor\_Latn} & Yoruba \\
\texttt{zul\_Latn} & Zulu & \texttt{sdh\_Arab} & Southern Kurdish \\
\texttt{smo\_Latn} & Samoan & \texttt{cos\_Latn} & Corsican \\
\texttt{hif\_Latn} & Fiji Hindi & \texttt{sun\_Latn} & Sundanese \\
\texttt{nya\_Latn} & Chichewa & \texttt{ibo\_Latn} & Igbo \\
\texttt{haw\_Latn} & Hawaiian & \texttt{sot\_Latn} & Sesotho \\
\texttt{lus\_Latn} & Mizo & \texttt{sna\_Latn} & Shona \\
\texttt{chv\_Cyrl} & Chuvash & \texttt{azb\_Arab} & South Azerbaijani \\
\texttt{roh\_Latn} & Romansh & \texttt{run\_Latn} & Kirundi \\
\texttt{hin\_Latn} & Hindi (Latin) & \texttt{sah\_Cyrl} & Yakut \\
\texttt{tir\_Ethi} & Tigrinya & \texttt{ast\_Latn} & Asturian \\
\texttt{oci\_Latn} & Occitan & \texttt{cnh\_Latn} & Hakha Chin \\
\texttt{sme\_Latn} & Northern Sami & \texttt{nds\_Latn} & Low German \\
\texttt{oss\_Cyrl} & Ossetian & \texttt{urd\_Latn} & Urdu (Latin) \\
\texttt{gsw\_Latn} & Swiss German & \texttt{anp\_Deva} & Angika \\
\texttt{apc\_Arab} & Levantine Arabic & \texttt{gaz\_Latn} & Oromo \\
\texttt{nap\_Latn} & Neapolitan & \texttt{hsb\_Latn} & Upper Sorbian \\
\texttt{hil\_Latn} & Hiligaynon & \texttt{kal\_Latn} & Greenlandic \\
\end{longtable}
\endgroup


\begin{thebibliography}{99}
\setlength{\itemsep}{0pt}
\bibitem{qwen2026qwen35}
Qwen Team. Qwen3.5: Towards Native Multimodal Agents. February 2026. \url{https://qwen.ai/blog?id=qwen3.5}.

\bibitem{shen2025multiway}
Yingli Shen, Wen Lai, Shuo Wang, Ge Gao, Kangyang Luo, Alexander Fraser, and Maosong Sun. From Unaligned to Aligned: Scaling Multilingual LLMs with Multi-Way Parallel Corpora. \emph{Proceedings of EMNLP}, pages 7357--7379, 2025. \url{https://aclanthology.org/2025.emnlp-main.374/}.

\bibitem{minicpm}
Shengding Hu et al. MiniCPM: Unveiling the Potential of Small Language Models with Scalable Training Strategies. \emph{arXiv:2404.06395}, 2024. \url{https://arxiv.org/abs/2404.06395}.

\bibitem{rival2025}
Tianjiao Li, Mengran Yu, Chenyu Shi, Yanjun Zhao, Xiaojing Liu, Qiang Zhang, Qi Zhang, Xuanjing Huang, and Jiayin Wang. RIVAL: Reinforcement Learning with Iterative and Adversarial Optimization for Machine Translation. \emph{arXiv:2506.05070}, 2025. \url{https://arxiv.org/abs/2506.05070}.

\bibitem{wortsman2022}
Mitchell Wortsman et al. Model soups: averaging weights of multiple fine-tuned models improves accuracy without increasing inference time. \emph{ICML}, PMLR 162:23965--23998, 2022. \url{https://proceedings.mlr.press/v162/wortsman22a.html}.

\bibitem{ma2026mopd}
Wenhan Ma, Jianyu Wei, Liang Zhao, Hailin Zhang, Bangjun Xiao, Lei Li, Qibin Yang, Bofei Gao, Yudong Wang, Rang Li, Jinhao Dong, Zhifang Sui, and Fuli Luo. MOPD: Multi-Teacher On-Policy Distillation for Capability Integration in LLM Post-Training. \emph{arXiv:2606.30406}, 2026. \url{https://arxiv.org/abs/2606.30406}.

\bibitem{xu2025qwen3omni}
Jin Xu et al. Qwen3-Omni Technical Report. \emph{arXiv:2509.17765}, 2025. \url{https://arxiv.org/abs/2509.17765}.

\bibitem{du2025cosyvoice3}
Zhihao Du et al. CosyVoice 3: Towards In-the-wild Speech Generation via Scaling-up and Post-training. \emph{arXiv:2505.17589}, 2025. \url{https://arxiv.org/abs/2505.17589}.

\bibitem{homura}
Ziang Cui et al. HOMURA: Taming the Sand-Glass for Time-Constrained LLM Translation via Reinforcement Learning. \emph{arXiv:2601.10187}, 2026. \url{https://arxiv.org/abs/2601.10187}.

\bibitem{kazemnejad2023nope}
Amirhossein Kazemnejad, Inkit Padhi, Karthikeyan Natesan Ramamurthy, Payel Das, and Siva Reddy. The Impact of Positional Encoding on Length Generalization in Transformers. \emph{NeurIPS}, 2023. \url{https://arxiv.org/abs/2305.19466}.

\bibitem{su2021roformer}
Jianlin Su, Yu Lu, Shengfeng Pan, Ahmed Murtadha, Bo Wen, and Yunfeng Liu. RoFormer: Enhanced Transformer with Rotary Position Embedding. \emph{arXiv:2104.09864}, 2021. \url{https://arxiv.org/abs/2104.09864}.

\bibitem{longwriter}
Yushi Bai et al. LongWriter: Unleashing 10,000+ Word Generation from Long Context LLMs. \emph{ICLR}, 2025. \url{https://arxiv.org/abs/2408.07055}.

\bibitem{nllb2022}
NLLB Team et al. No Language Left Behind: Scaling Human-Centered Machine Translation. \emph{arXiv:2207.04672}, 2022. \url{https://arxiv.org/abs/2207.04672}.

\bibitem{deutsch2025wmt24pp}
Daniel Deutsch et al. WMT24++: Expanding the Language Coverage of WMT24 to 55 Languages \& Dialects. \emph{Findings of ACL}, pages 12257--12284, 2025. \url{https://aclanthology.org/2025.findings-acl.634/}.

\bibitem{zheng2026hymt2}
Mao Zheng et al. Hy-MT2: A Family of Fast, Efficient and Powerful Multilingual Translation Models in the Wild. \emph{arXiv:2605.22064}, 2026. \url{https://arxiv.org/abs/2605.22064}.

\bibitem{wang2023guofeng}
Longyue Wang et al. Findings of the WMT 2023 Shared Task on Discourse-Level Literary Translation: A Fresh Orb in the Cosmos of LLMs. \emph{Proceedings of the Eighth Conference on Machine Translation}, pages 55--67, 2023. \url{https://aclanthology.org/2023.wmt-1.3/}.

\bibitem{rei2022comet22}
Ricardo Rei et al. COMET-22: Unbabel-IST 2022 Submission for the Metrics Shared Task. \emph{Proceedings of the Seventh Conference on Machine Translation}, pages 578--585, 2022. \url{https://aclanthology.org/2022.wmt-1.52/}.

\bibitem{li2026mmtitbench}
Gengluo Li, Chengquan Zhang, Yupu Liang, Huawen Shen, Yaping Zhang, Pengyuan Lyu, Weinong Wang, Xingyu Wan, Gangyan Zeng, Han Hu, Can Ma, and Yu Zhou. MMTIT-Bench: A Multilingual and Multi-Scenario Benchmark with Cognition-Perception-Reasoning Guided Text-Image Machine Translation. \emph{Proceedings of the IEEE/CVF Conference on Computer Vision and Pattern Recognition}, pages 16593--16602, 2026. \url{https://openaccess.thecvf.com/content/CVPR2026/html/Li_MMTIT-Bench_A_Multilingual_and_Multi-Scenario_Benchmark_with_Cognition-Perception-Reasoning_Guided_Text-Image_CVPR_2026_paper.html}.

\bibitem{salesky2024vistra}
Elizabeth Salesky, Philipp Koehn, and Matt Post. Benchmarking Visually-Situated Translation of Text in Natural Images. \emph{Proceedings of the Ninth Conference on Machine Translation}, pages 1167--1182, 2024. \url{https://aclanthology.org/2024.wmt-1.115/}.

\bibitem{guerreiro2024xcomet}
Nuno M. Guerreiro, Ricardo Rei, Daan van Stigt, Luisa Coheur, Pierre Colombo, and Andr\'e F. T. Martins. xCOMET: Transparent Machine Translation Evaluation through Fine-grained Error Detection. \emph{Transactions of the Association for Computational Linguistics}, 12:979--995, 2024. \url{https://aclanthology.org/2024.tacl-1.54/}.

\bibitem{huang2023ceval}
Yuzhen Huang et al. C-Eval: A Multi-Level Multi-Discipline Chinese Evaluation Suite for Foundation Models. \emph{NeurIPS}, 2023. \url{https://arxiv.org/abs/2305.08322}.

\bibitem{rein2023gpqa}
David Rein et al. GPQA: A Graduate-Level Google-Proof Q\&A Benchmark. \emph{arXiv:2311.12022}, 2023. \url{https://arxiv.org/abs/2311.12022}.

\bibitem{romanou2024include}
Angelika Romanou et al. INCLUDE: Evaluating Multilingual Language Understanding with Regional Knowledge. \emph{arXiv:2411.19799}, 2024. \url{https://arxiv.org/abs/2411.19799}.

\bibitem{openai2024mmmlu}
OpenAI. Multilingual Massive Multitask Language Understanding (MMMLU). Dataset. \url{https://huggingface.co/datasets/openai/MMMLU}.

\bibitem{jiang2022bwb}
Yuchen Jiang et al. BlonDe: An Automatic Evaluation Metric for Document-level Machine Translation. \emph{NAACL-HLT}, pages 1550--1565, 2022. \url{https://aclanthology.org/2022.naacl-main.111/}.

\bibitem{wang2025segale}
Kuang-Da Wang, Shuoyang Ding, Chao-Han Huck Yang, Ping-Chun Hsieh, Wen-Chih Peng, Vitaly Lavrukhin, and Boris Ginsburg. Extending Automatic Machine Translation Evaluation to Book-Length Documents. \emph{EMNLP}, pages 32323--32339, 2025. \url{https://aclanthology.org/2025.emnlp-main.1645/}. Code: \url{https://github.com/NVlabs/SEGALE}.

\bibitem{rei2020comet}
Ricardo Rei, Craig Stewart, Ana C Farinha, and Alon Lavie. COMET: A Neural Framework for MT Evaluation. \emph{EMNLP}, pages 2685--2702, 2020. \url{https://aclanthology.org/2020.emnlp-main.213/}.

\bibitem{gao2023reward}
Leo Gao, John Schulman, and Jacob Hilton. Scaling Laws for Reward Model Overoptimization. \emph{ICML}, PMLR 202:10835--10866, 2023. \url{https://proceedings.mlr.press/v202/gao23h.html}.

\bibitem{raina2024judge}
Vyas Raina, Adian Liusie, and Mark Gales. Is LLM-as-a-Judge Robust? Investigating Universal Adversarial Attacks on Zero-shot LLM Assessment. \emph{EMNLP}, pages 7499--7517, 2024. \url{https://aclanthology.org/2024.emnlp-main.427/}.
\bibitem{translatebooks}
Hydropix. TranslateBooksWithLLMs. Open-source book translation framework, version 1.5.13. \url{https://github.com/hydropix/TranslateBooksWithLLMs}.

\bibitem{gemini38flash}
Google. Gemini 3.8 Flash model documentation. Accessed September 29, 2026. \url{https://ai.google.dev/gemini-api/docs/models/gemini-3.8-flash}.

\end{thebibliography}
\end{document}